\documentclass[10pt,twocolumn,letterpaper]{article}

 \usepackage[pagenumbers]{cvpr} % To force page numbers, e.g. for an arXiv version

\usepackage{colortbl}
\usepackage{pifont} % add this in the preamble
\newcommand{\cmark}{\ding{51}} % checkmark
\newcommand{\xmark}{\ding{55}} % xmark

\definecolor{LightGray}{rgb}{0.9,0.9,0.9}
\definecolor{DarkRed}{rgb}{0.75,0,0}
\definecolor{DarkBlue}{rgb}{0,0,0.55}
\definecolor{DarkGreen}{rgb}{0.43, 0.68, 0.28}

\definecolor{xishen}{rgb}{0.858, 0.188, 0.478}

\definecolor{DarkBlue}{rgb}{0.0, 0.5, 0.8}
\definecolor{cvprblue}{rgb}{0.21,0.49,0.74}
\usepackage[pagebackref,breaklinks,colorlinks,allcolors=cvprblue]{hyperref}

\usepackage[accsupp]{axessibility} % Improves PDF readability for those with disabilities.

\def\paperID{38018} % *** Enter the Paper ID here
\def\confName{CVPR}
\def\confYear{2026}

\title{Learning Cross-View Object Correspondence via Cycle-Consistent Mask Prediction}

\author{
    \textbf{Shannan Yan}$^{1,2}$  \quad
    \textbf{Leqi Zheng}$^{1}$  \quad
    \textbf{Keyu Lv}$^{1}$ \quad
    \textbf{Jingchen Ni}$^{1}$  \quad
    \textbf{Hongyang Wei}$^{1}$  \\
    \textbf{Jiajun Zhang}$^{3}$  \quad
    \textbf{Guangting Wang}$^{2}$  \quad
    \textbf{Jing LYU}$^{2}$ \quad
    \textbf{Chun Yuan}$^{1\dagger}$ \quad
    \textbf{Fengyun Rao}$^{2\dagger}$ \quad
    \\[.6em]
    $^1$Tsinghua University \quad
    $^2$WeChat Vision, Tencent Inc. \quad
    $^3$USTC 
    \\[.5em]
    {\tt\small 
        ysn24@mails.tsinghua.edu.cn, yuanc@sz.tsinghua.edu.cn, fengyunrao@tencent.com
    }
}

\begin{document}

\maketitle

\newcommand\blfootnote[1]{%
  \begingroup
  \renewcommand\thefootnote{}\footnote{#1}%
  \addtocounter{footnote}{-1}%
  \endgroup
}
\blfootnote{$^\dagger$Corresponding author. This work was done when Shannan Yan was an intern at Tencent Inc.} 

\vspace{-1.5em}
\begin{abstract}
   We study the task of establishing object-level visual correspondence across different viewpoints in videos, focusing on the challenging egocentric-to-exocentric and exocentric-to-egocentric scenarios. We propose a simple yet effective framework based on conditional binary segmentation, where an object query mask is encoded into a latent representation to guide the localization of the corresponding object in a target video. To encourage robust, view-invariant representations, we introduce a cycle-consistency training objective: the predicted mask in the target view is projected back to the source view to reconstruct the original query mask. This bidirectional constraint provides a strong self-supervisory signal without requiring ground-truth annotations and enables test-time training (TTT) at inference. Experiments on the Ego-Exo4D and HANDAL-X benchmarks demonstrate the effectiveness of our optimization objective and TTT strategy, achieving state-of-the-art performance. The code is available at \url{https://github.com/shannany0606/CCMP}.
\end{abstract}   
\vspace{-1.5em}
\section{Introduction}
\label{sec:intro}

Understanding visual correspondence across different viewpoints is a core capability for embodied agents operating in complex environments. In applications such as human-robot interaction~\cite{alam2022social,bartneck2024human}, autonomous navigation~\cite{turan2022autonomous,zhang2022autonomous}, and assistive robotics~\cite{kyrarini2021survey,pang2022skin}, an agent must consistently identify and reason about the same object or scene across drastically different visual perspectives. For instance, a service robot may need to interpret instructions from an egocentric wearable camera and then locate the referred object from its own third-person viewpoint. Achieving such cross-view correspondence is crucial for grounding language, enabling coordination, and executing goal-directed actions. However, the substantial viewpoint disparity between egocentric (first-person) and exocentric (third-person) cameras introduces major challenges such as appearance variation, occlusions, and disjoint spatial references, which traditional correspondence models, often trained on co-visible or static scenes, fail to handle effectively.

\begin{figure}
\centering
\vspace{-0.1in}
\includegraphics[width=0.48\textwidth]{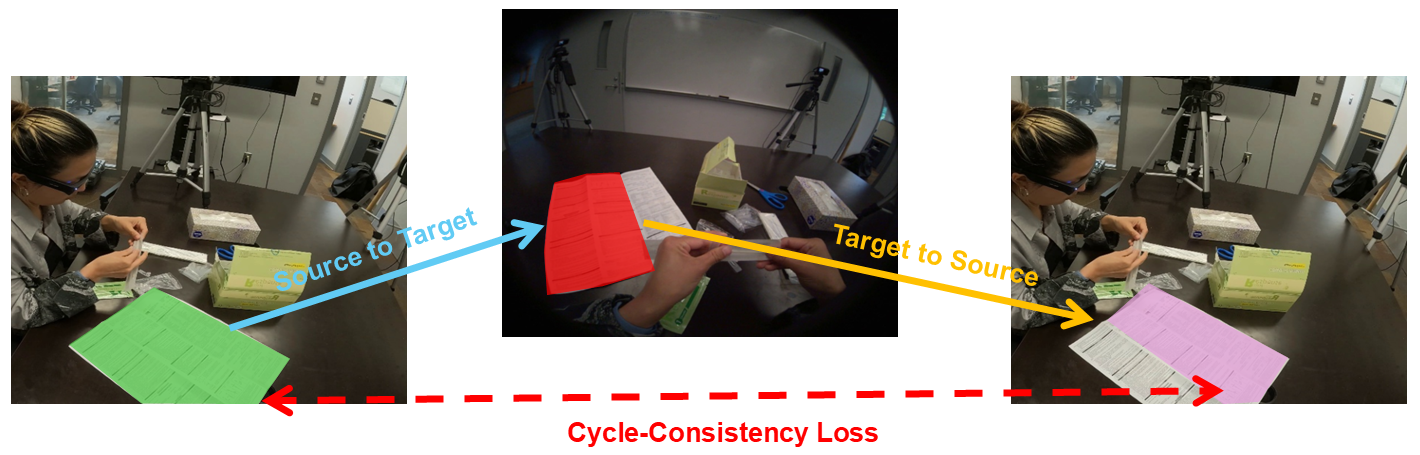}
\vspace{-0.25in}
\caption{
{\textbf{Cycle-Consistent Visual Correspondence with Test-Time Training.}
Our framework learns object-level correspondences by enforcing cycle consistency: the object mask is transferred from source to target view and projected back to reconstruct the original query. This self-supervised constraint enables robust cross-view alignment and supports test-time training to further improve performance during inference.} 
}
\vspace{-0.2in}
\label{fig:teaser}
\end{figure}

The task is challenging due to several compounding factors. First, the visual appearance of objects can vary drastically across views because of changes in camera angle, lighting, occlusion, and resolution. Egocentric views are often shaky, cluttered, and subject to motion blur, while exocentric views are more stable but may lack fine-grained detail. Second, the spatial layout and context around an object can differ significantly across viewpoints, making it difficult to rely on background cues for matching. Third, establishing correspondence requires reasoning not just over spatial features but also over temporal dynamics, as objects may move or deform differently across camera views. These challenges make traditional appearance-based or tracking-based methods insufficient and call for models that can learn robust, view-invariant object representations while being sensitive to fine-grained semantic and temporal cues.

%In this paper, we focus on the problem of cross-view object-level visual correspondence in videos, where the goal is to localize the same object instance across egocentric and exocentric perspectives. This setting is especially relevant for scenarios such as human-robot collaboration, wearable camera systems, and surveillance, where understanding the spatial and semantic alignment between disparate viewpoints is crucial.

 To address the challenge of establishing cross-view visual correspondence, we propose a simple yet effective framework based on conditional binary segmentation. Our approach leverages the powerful vision foundation model 
 DINOv3~\cite{oquab2023dinov2} as backbone, and introduces a single conditioning token (\(\mathit{CDT}\)) to inject source image information into the vision transformer. The modified visual tokens are then used to predict a binary mask in the target view. This compact design enables cross-view alignment with minimal architectural changes and maintains full compatibility with pretrained backbones.

To enhance training supervision, we introduce a cycle-consistency objective that enforces consistency between the original query mask and its round-trip projection through the target view. As illustrated in Figure~\ref{fig:teaser}, the predicted mask originates in the source view, is generated in the target view, and is expected to reconstruct the initial query mask. This bidirectional constraint encourages the model to learn view-invariant representations and provides self-supervision in the absence of target-view annotations. Crucially, it also enables test-time training (TTT) at inference time, further improving correspondence quality under domain shift or distributional variation.

%Instead of directly regressing a binary mask, we compute a dense similarity map between the query token and the image tokens of the target frame using normalized feature distances. This explicit similarity computation provides better interpretability and higher mask quality. We further enhance learning stability and fine-grained alignment by incorporating an auxiliary loss that encourages the model to maintain strong associations between matching regions across views. Training proceeds in two stages. In the first stage, we optimize the model with a standard segmentation loss (e.g., mIoU) to learn fine-grained spatial correspondence. In the second stage, to assess object visibility—a coarser but semantically important task—we freeze the network parameters and train a lightweight classification head on top of the global [CLS] token. This staged training strategy eases optimization and leads to improved performance for visibility prediction. To further promote robust and view-invariant representations, we introduce a cycle-consistency training objective. Specifically, the predicted mask in the target view is projected back to the source frame and used to reconstruct the original query mask. This forward-backward consistency constraint enforces semantic alignment without requiring additional supervision or paired labels, and encourages the model to learn transferable features that generalize across diverse camera viewpoints and visual contexts.

We validate our approach on the challenging Ego-Exo4D~\cite{grauman2024ego} dataset, which contains diverse egocentric and exocentric video pairs with rich object-level annotations. In the Exo Query setup, our method outperforms all prior baselines by over 3.10\%, demonstrating strong cross-view correspondence capabilities. In the Ego Query setting, we achieve an IoU of 41.95\%, closely approaching the previous state-of-the-art (SOTA) method O-MaMa~\cite{mur2025mama}, which scores 42.57\%. We further evaluate on HANDAL-X~\cite{fu2024objectrelator}, where our method outperforms ObjectRelator by 36.0\% on zero-shot segmentation. Incorporating the cycle-consistency constraint during training and leveraging TTT at inference consistently yield improvements across both setups. Despite its conceptual simplicity, our framework effectively captures fine-grained correspondences under extreme viewpoint changes.

In summary, our contributions are:
\begin{itemize}[leftmargin=2em]
\item We propose a simple, modular, and end-to-end framework for cross-view visual object correspondence that leverages vision foundation models with minimal architectural modifications.
\item We introduce a novel cycle-consistency objective that enforces semantic alignment between source and target views. This self-supervisory signal enables effective test-time training during inference.
\item We validate our approach on the Ego-Exo4D and HANDAL-X benchmarks, achieving SOTA performance and demonstrating its strong effectiveness.
\end{itemize}

\vspace{-.5em}
\section{Related Works}

\paragraph{Cross-view Video Understanding.}
Bridging ego- and exo-centric perspectives can enrich video understanding, yet most work targets one view. Egocentric research spans classification~\cite{liu2022video,woo2023convnext}, question answering~\cite{zhang2023video,alayrac2022flamingo}, and captioning~\cite{yang2023vid2seq,lin2022swinbert}, but lagged behind exocentric methods due to scarce data. The advent of large-scale benchmarks like Ego4D~\cite{grauman2022ego4d}, EPIC-KITCHENS-100~\cite{damen2022rescaling}, and richer modeling~\cite{jia2022egotaskqa,liu2022joint} has closed this gap. However, few works connect both views~\cite{luo2024put,liu2024exocentric}, until Ego-Exo4D’s time-aligned annotations~\cite{grauman2024ego}. Recently, Baade et al.~\cite{baade2025self} synthesize paired masks from raw segmentations using predictive cycle consistency and iterative pseudo-labeling. ObjectRelator~\cite{fu2024objectrelator} fine-tunes PSALM~\cite{zhang2024psalm} with auxiliary modules that enforce view-invariant embeddings through self-supervised alignment. O-MaMa~\cite{mur2025mama} reformulates cross-view segmentation as a mask-matching problem by integrating FastSAM~\cite{zhao2023fast} to generate candidate masks in advance. In this paper, we present an end-to-end baseline that requires no extra data and does not rely on auxiliary modules.

\paragraph{Vision Foundation Models.} 

With the rapid advancement of deep learning and large language models~\cite{zheng2025lagcl4rec,yu2025comrope,zheng2025negative,yan2023redualsvg,yan2026adamem,zheng2026should,kang2025hssbench,lv2026makes,su2026generation,wang2025iterprime,ni2025semantic,ni2026fcl,SRI,luo2026prompthub}, 
recent progress in vision foundation models has significantly advanced representation learning by leveraging transformer architectures and self-supervised techniques. Vision Transformer (ViT)~\cite{dosovitskiy2020image} introduced a transformer-based approach for image recognition, followed by DeiT~\cite{touvron2021training} and DeiT3~\cite{touvron2022deit}, which improved training efficiency and performance on limited data. MoCo v3~\cite{chen2021empirical} extended contrastive learning to ViTs, while MAE~\cite{he2022masked} proposed masked image reconstruction as a pretext task. DINO~\cite{caron2021emerging}, DINOv2~\cite {oquab2023dinov2} and DINOv3~\cite{simeoni2025dinov3} employed self-distillation to learn rich semantic features. CLIP~\cite{radford2021learning} aligned vision and language via contrastive learning on image-text pairs, inspiring extensions like SigLIP~\cite{zhai2023sigmoid} and SigLIP2~\cite{tschannen2025siglip}, which replaced contrastive loss with sigmoid-based objectives for better cross-modal alignment. These models form the foundation of modern vision systems, demonstrating strong generalization and scalability across diverse tasks. However, none are explicitly designed for ego-exo correspondence, which remains particularly challenging.

\paragraph{Test-Time Training.} 

Test-time training (TTT) has evolved from a self-supervised adaptation method for distribution shifts into a versatile framework across modalities, including images, videos, and language. Early work~\cite{sun19ttt} proposed optimizing a model on each unlabeled test sample via a self-supervised task to enhance robustness under covariate shifts. Subsequent studies~\cite{gandelsman2022test, wang2025test, hardt2024test, sun2024learning, dalal2025one} expanded TTT to masked autoencoding, streaming video adaptation, retrieval-augmented language modeling, and generative video Transformers, demonstrating its broad applicability and growing impact. To the best of our knowledge, our method is the first to successfully apply TTT to this task and achieve a clear performance improvement.

% Test-time training (TTT) has rapidly evolved from its inception as a self-supervised adaptation technique for handling distribution shifts to a versatile framework across modalities—including images, videos, language, and sequence modeling. Early work~\cite{sun19ttt} introduced the core idea of optimizing a model on each unlabeled test sample via a self-supervised auxiliary task to improve robustness under covariate shifts. Gandelsman et al.~\cite{gandelsman2022test} demonstrated the power of masked autoencoders to provide strong self-supervised objectives for one-sample adaptation, significantly boosting performance on corrupted vision benchmarks. Building on these foundations, recent research~\cite{wang2025test, hardt2024test, sun2024learning} has extended TTT to streaming video data, retrieval‑augmented language modeling, and novel sequence modeling layers that redefine hidden states as learnable functions. The latest work~\cite{dalal2025one} applies TTT layers to generative video Transformers, enabling coherent one‑minute video synthesis—a testament to TTT’s broad applicability and continuing impact.
\vspace{-.5em}
\section{Approach}
\label{sec:method}

In this section, we present our approach to visual correspondence. Given a source image $I_s$, a target image $I_t$, and an object mask $M_s$ in $I_s$, the objective is to accurately segment the corresponding object $M_t$ in the target image. To ensure clarity and consistency, we define \textbf{Ego2Exo} as the task where the ego-centric view (circular field of view) serves as the query and the exo-centric view as the target, and \textbf{Exo2Ego} as the reverse setting.

The section is organized as follows: Section~\ref{subsec:pipeline} introduces our simple yet effective pipeline, Section~\ref{subsec:loss} details the objective function, and the implementation details are provided in Section~\ref{sec:imple_details}.

\begin{figure*}
\centering
\includegraphics[width=1.0\textwidth]{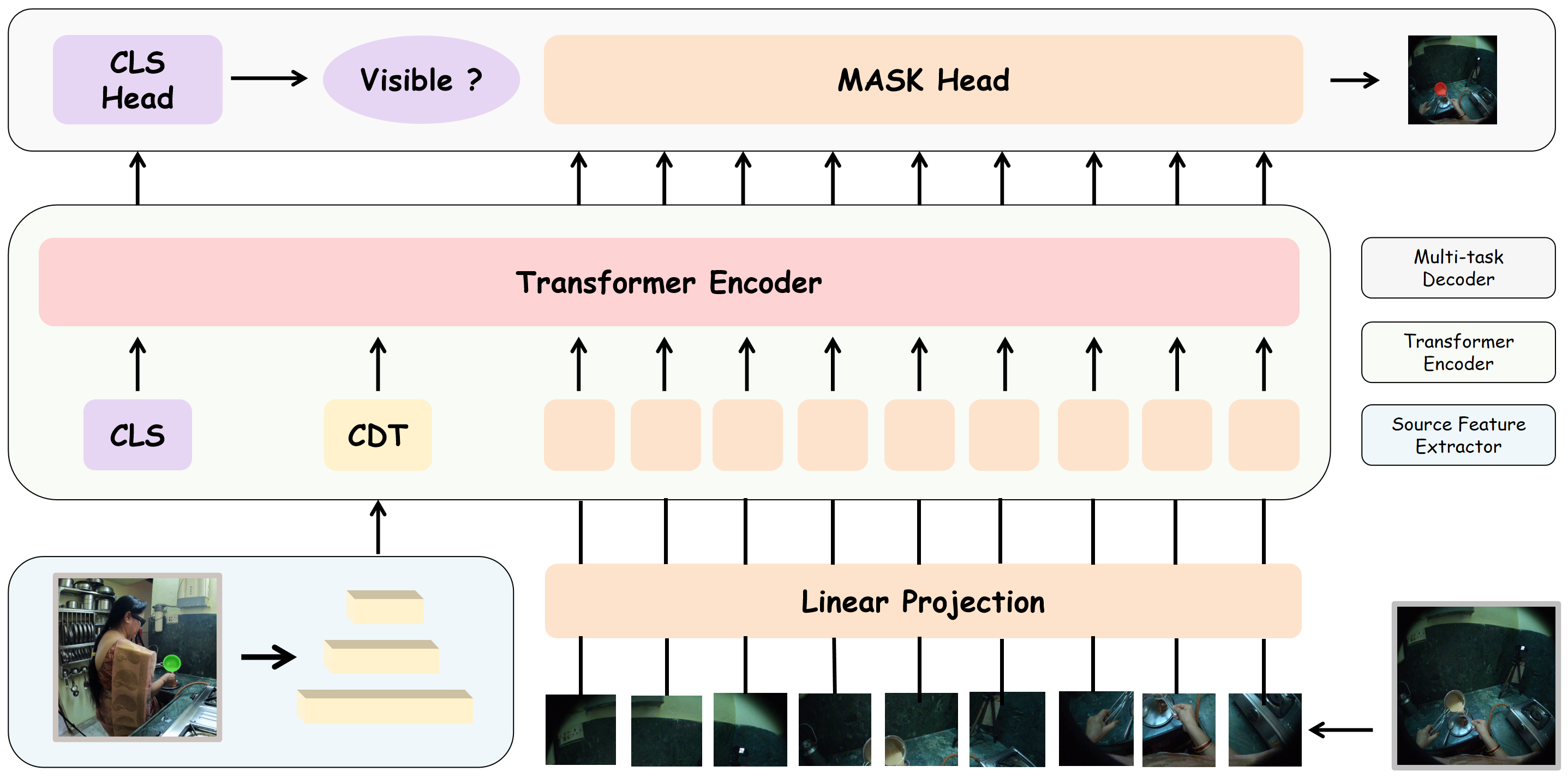}
\vspace{-0.15in}
\caption{
\textbf{Model overview.} \( \mathit{CLS} \) denotes class tokens, and \( \mathit{CDT} \) denotes condition tokens. The CLS head determines whether the object in the target image corresponding to a given object mask in the source image is visible. The bottom-left image shows the source image with the object mask, while the bottom-right image shows the target image.
}
\vspace{-0.05in}
\label{fig:overview}
\end{figure*}

\subsection{Pipeline Overview}
\label{subsec:pipeline}
We propose a transformer-based framework that leverages conditional features and a mask-guided attention mechanism to establish robust visual correspondences.

The overall pipeline is illustrated in Figure~\ref{fig:overview}. Our model comprises three main components: Source Feature Extractor, Transformer Encoder, and Multi-task Decoder.

\paragraph{Source Feature Extractor.}  
The objective of this module is to extract an object-specific feature representation from a source image \( I_s \) using its corresponding mask \( M_s \). First, we obtain the feature map \( F_s \in \mathbb{R}^{C \times H \times W} \) from \( I_s \) using a backbone network \( \mathcal{F}_{sfe}(\cdot) \): $F_s = \mathcal{F}_{sfe}(I_s)$.

The mask \( M_s \) is resized (if necessary) to match the spatial dimensions of \( F_s \) and is then normalized such that its elements sum to one. Specifically, we compute the normalized mask \( \tilde{M}_s \) as: $\tilde{M}_s = \frac{M_s}{\sum_{i,j} M_s[i,j] + \tau}$, where \( \tau \) is a small hyperparameter (typically \( 1 \times 10^{-6} \)) used to prevent numerical errors. This yields a compact, stable, and scale-invariant object-focused representation. 

Next, we compute the masked object feature \( z_s \in \mathbb{R}^{C} \) as a weighted average of the feature map \( F_s \) over the spatial locations, using \( \tilde{M}_s \) as the weights:$z_s = \sum_{i=1}^{H} \sum_{j=1}^{W} \tilde{M}_s[i, j] \cdot F_s[:, i, j]$.

This produces a compact representation that highlights the regions specified by the mask, which is then projected onto the condition token \( \mathit{CDT} \) in the transformer encoder.

\paragraph{Transformer Encoder.} The target image \( I_t \) is divided into patches, which are then linearly projected to form n visual tokens \( [x_1, x_2, \dots, x_n] \) for the transformer encoder. Along with the condition token \( \mathit{CDT} \) and the class token \( \mathit{CLS} \), the final input to the transformer encoder is
$$\mathbf{x}_{\text{input}} = [\mathit{CLS}, \mathit{CDT}, x_1, x_2, \dots, x_n].$$ 
These tokens are fed into a standard transformer encoder, enabling the \( \mathit{CDT} \) to condition the transformer features through cross-token attention, which facilitates object-aware representation learning in the target image.

\paragraph{Multi-task Decoder.} The output tokens from the transformer encoder are processed by two parallel heads: the \textbf{Mask Head}, which generates the feature for each visual token \( y_i \);  and the \textbf{CLS Head}, which, with an additional classification token \( \mathit{CLS} \), predicts whether the object in \( I_s \) is visible in \( I_t \) (i.e., performs binary visibility classification).

% The final segmentation mask \( \hat{M}_t \) is generated using both the visual tokens and the updated condition token \( y_{\mathit{cdt}} \). Specifically, for the \( i \)-th token \( y_i \), the mask prediction follows the formulation presented in Eq.~\ref{sigmoid_loss}, which is inspired by the sigmoid loss~\cite{zhai2023sigmoid}.

% \begin{equation}
% \label{sigmoid_loss}
%     \hat{M}^i_t = \mathtt{Sigmoid} (\tau \mathtt{Cos}(y_{\mathit{cdt}}, y_i) - \beta )
% \end{equation}

%  where $\mathtt{Cos}$ is the cosine similarity: $\mathtt{Cos}(a, b) = \frac{a \cdot b}{\|a\| \|b\|}$,  $\beta$ and $\tau$ are learnable bias and temperature similar to~\cite{zhai2023sigmoid}. Note that the final mask is obtained by applying bilinear upsampling to the predicted mask \( \hat{M}_t \) to match the target resolution. For simplicity, we omit the distinction and refer to \( \hat{M}_t \) directly as the final mask prediction in the following discussion.

The final segmentation mask \( \hat{M}_t \) is generated employing a lightweight prediction head consisting of two convolutional layers applied solely on the visual tokens.  

The overall design allows the model to jointly reason about spatial alignment, object visibility, and semantic consistency, resulting in improved correspondence performance under complex appearance and viewpoint variations.

% Intuitively, the mask prediction mechanism enables the model to adaptively generate segmentation masks conditioned on the provided object features. This overall design allows the model to jointly reason about spatial alignment, object visibility, and semantic consistency, resulting in improved correspondence performance under complex appearance and viewpoint variations.

\subsection{Learning Visual Correspondences}

\subsubsection{Objective Function}
\label{subsec:loss}

We employ multiple objective functions to effectively learn visual correspondences: the mask loss \( \mathcal{L}_{\text{mask}} \), the auxiliary loss \( \mathcal{L}_{\text{aux}} \), and the cycle-consistency loss \( \mathcal{L}_{\text{cycle}} \). The total training objective is defined as

\begin{equation}
\label{total_loss}
    \mathcal{L}_{\text{total}} = \mathcal{L}_{\text{mask}} + \lambda_{\text{aux}} \mathcal{L}_{\text{aux}} + \lambda_{\text{cycle}} \mathcal{L}_{\text{cycle}},
\end{equation}
where \( \lambda_{\text{aux}} \) and \( \lambda_{\text{cycle}} \) are hyperparameters that balance the contributions of the auxiliary and cycle-consistency losses, respectively.

\paragraph{Mask Loss.} Since our task is binary segmentation, we adopt a combination of Binary Cross-Entropy (BCE) loss and Dice loss to supervise the predicted target mask $\hat M_t$, which is similar to~\cite{grauman2024ego}. Precisely, our mask loss is composed of $\mathcal{L}_{\text{bce}}$ as well as $\mathcal{L}_{\text{dice}}$. 

\begin{equation}
\label{mask_loss}
\begin{aligned}
& \mathcal{L}_{\text{mask}}(M_t, \hat M_t) = \mathcal{L}_{\text{bce}}(M_t, \hat M_t) + \lambda_{\text{dice}} \mathcal{L}_{\text{dice}}(M_t, \hat M_t), \\
\end{aligned}
\end{equation}
where $\mathcal{L}_{\text{bce}}(M_t, \hat M_t)$ and $\mathcal{L}_{\text{dice}}(M_t, \hat M_t)$ are defined as
\begin{equation}
\label{eq3}
\begin{aligned}
& \mathcal{L}_{\text{bce}}(M_t, \hat M_t) = \\ & \ \ \ \ \ \ \ - \frac{1}{N} \sum_i \left[ M^i_t \log(\hat M^i_t) + (1 - M^i_t) \log(1 - \hat M^i_t) \right], \\
& \mathcal{L}_{\text{dice}}(M_t, \hat M_t) = 1 - \frac{2 \sum_{i} M^i_t \hat M^i_t + \epsilon}{\sum_i M^i_t + \sum_i \hat M^i_t + \epsilon} \ ,\\
\end{aligned}
\end{equation}
where $\epsilon$ is a small constant (e.g., $10^{-6}$ to prevent division by zero).  % We find that optimizing \( \mathcal{L}_{\text{dice}} \) is crucial, as the target objects often occupy a very small area in the mask. Supporting experiments are provided in Section~\todo{XXX, if we cannot provide experiments, provide them in the supplementary material.}.

\paragraph{Auxiliary Loss.} 
To facilitate training and improve gradient flow, we introduce an auxiliary supervision signal by applying the same mask loss to intermediate predictions from the last few transformer encoder layers. Specifically, the auxiliary loss is computed between the ground-truth target mask \( M_t \) and the intermediate predicted masks, using the same formulation as in Eq.~\ref{mask_loss}. The total auxiliary loss is averaged across selected layers. This deep supervision encourages the model to learn meaningful representations at different levels of the network.

\paragraph{Cycle-consistency Loss.}
To improve the robustness of learned visual correspondences, we introduce a cycle-consistency constraint. The key idea is to map the source mask $M_s$ to the predicted target mask $\hat{M}_t$, and then map it back to the reconstructed source mask $\hat{M}_s$, ensuring that $\hat{M}_s$ closely matches the original $M_s$. Formally, the cycle-consistency loss is defined as:

\begin{equation}
\label{cycle_loss}
\mathcal{L}_{\text{cycle}} = \mathcal{L}_{\text{bce}}(M_s, \hat{M}_s),
\end{equation}
where $\mathcal{L}_{\text{bce}}$ is the part of binary cross entropy loss used in Eq.~\ref{mask_loss}. This objective encourages the model to learn more consistent and reliable correspondences by enforcing a closed-loop mapping between source and target domains.

Importantly, the cycle-consistency loss $\mathcal{L}_{\text{cycle}}$ does not rely on ground-truth target masks, making it applicable during inference for test-time training (TTT). We do not explicitly handle the corner case of invisible objects in the cycle, as such instances are rare in Ego-Exo4D.

\subsubsection{Visibility Prediction and Test-Time Training}

\paragraph{Visibility Prediction.}  
Since visibility prediction is inherently an instance-level task that determines whether an object is visible as a whole rather than a pixel-level segmentation task, we treat it separately from mask prediction. To this end, we introduce a lightweight post-training step specifically for visibility classification. After training the main model, we freeze the entire network and fine-tune only the \( \mathit{CLS} \) Head, which serves as a binary classifier applied to the \( \mathit{CLS} \) token. This design allows us to leverage the instance-level semantic representation encoded in the \( \mathit{CLS} \) token without altering the learned correspondence or segmentation capabilities of the backbone.

\paragraph{Test-Time Training.}  
Since the cycle-consistency loss \( \mathcal{L}_{\text{cycle}} \) does not require ground-truth target masks, it can be leveraged at inference time to further refine the model. Specifically, we apply test-time training (TTT) for image pairs. During TTT, we fine-tune only the last \( K \) transformer encoder layers of the model for each test pair, using \( T \) gradient update steps with a learning rate of \( lr_{\text{ttt}} \). This allows the model to adapt to the specific test pair and improve correspondence quality. % We ablate the choice of \( K \), \( T \), and \( lr_{\text{ttt}} \) in the supplementary material.

\subsection{Implementation Details}
\label{sec:imple_details}

\paragraph{Architecture.}  
We adopt the ConvNeXt-based pretrained DINOv3-L model~\cite{simeoni2025dinov3} as the source feature extractor for its efficiency. For the transformer encoder, we employ the ViT-based pretrained DINOv3-L model, which provides rich visual representations. The compact feature representation obtained from the source image is linearly projected to form the \( \mathit{CDT} \) token, aligning the feature dimension with the transformer input. 

\paragraph{Training Details.}  
% We train our model using the AdamW optimizer~\cite{loshchilovdecoupled} with a cosine learning rate schedule and linear warm-up. We use a batch size of 16. The image size is 512×512. We use a batch size of 16.  
The training process on Ego-Exo4D consists of two stages. In the first stage (\textit{linear probing}), we freeze the two DINOv3 backbones and train the remaining modules for 64K iterations. % The learning rate decays from the maximum value of $1\times10^{-3}$ to a minimum of $1\times10^{-4}$. 
In the second stage, all parameters are unfrozen and optimized for 640K iterations. % The learning rate decays from the maximum value of $1\times10^{-5}$ to a minimum of $1\times10^{-6}$. 
To address GPU memory limitations (40GB), we adopt gradient accumulation with a step size of 16, resulting in an effective number of parameter updates of 704K / 16 = 44K. The training process takes approximately 72 hours on 8 NVIDIA RTX A800 GPUs. % We maintain an exponential moving average (EMA) of the model parameters throughout training, and use the EMA model as the final model for evaluation. On HANDAL-X, we train for 10 epochs with the learning rate decaying from the maximum value of $2\times10^{-4}$ to a minimum of $2\times10^{-6}$.
For visibility prediction, we fine-tune only the CLS Head for 96K iterations, which takes approximately 1 hour on the same hardware.  %, using the same training setup as the main binary segmentation task. 
More training details are provided in the supplementary material.

\paragraph{Hyperparameter Settings.}  
%In Eq.~\ref{sigmoid_loss}, the parameters $\tau$ and $\beta$ are initialized to 10 and 5, respectively. 
In the total loss formulation, we set the loss weights as follows: $\lambda_{\text{dice}} = 5$, $\lambda_{\text{aux}} = 1$, and $\lambda_{\text{cycle}} = 10$. The auxiliary loss is applied to the second-to-last layer of the transformer encoder. For TTT, we update the last $K{=}4$ layers for $T{=}2$ steps in Ego2Exo and the last $K{=}11$ layers for $T{=}6$ steps in Exo2Ego, using a learning rate of $5\times10^{-6}$.
\section{Experiments}
\label{sec:exp}

\subsection{Experimental Setting}
\label{sec:exp_setting}

\paragraph{Datasets.}
%To evaluate the effectiveness of our approach, we adopt the Ego-Exo4D correspondence dataset~\cite{grauman2024ego}. The dataset comprises 1.8 million annotated object masks sampled at 1 fps from 1,335 video takes, spanning diverse domains such as dance, soccer, basketball, bouldering, music, cooking, bike repair, and healthcare, all captured in natural, unscripted settings. The Ego-Exo4D’s object correspondence benchmark is unique among existing datasets in emphasizing highly challenging conditions: many small objects, severe occlusions, and extreme viewpoint differences in dynamic settings. However, since Ego-Exo4D redacted some details prior to final release due to some privacy restrictions, there are 66 takes in train split (821 takes in total) unavailable to us, making it harder for us to outperform previous methods.
To evaluate the effectiveness of our approach, we adopt the Ego-Exo4D correspondence dataset~\cite{grauman2024ego}. This dataset contains 1.8 million annotated object masks sampled at 1 fps from 1,335 video takes, covering a wide range of domains such as soccer, basketball, music, cooking, bike repair, and healthcare, all recorded in natural, unscripted environments. Due to privacy restrictions, 66 training takes (out of 821) were redacted from the released dataset, limiting our training data and making direct comparison with prior works less favorable. We thus use 755 takes for training, 201 for validation, and 295 for testing.

To further evaluate the generalization ability of different methods, we conduct experiments on HANDAL-X, a cross-view object segmentation benchmark introduced by ObjectRelator~\cite{fu2024objectrelator}. HANDAL-X contains multi-view image pairs that capture objects from complete 360° viewpoints with corresponding object-centric masks. The dataset comprises 44,102 training pairs and 14,074 test pairs. Unlike the egocentric–exocentric Ego-Exo4D, HANDAL-X provides a complementary setting for assessing cross-view correspondence under broader viewpoint variations.

\paragraph{Data Preprocessing.}
To enhance the diversity of training data, we adopt three complementary preprocessing strategies.
First, we unify the Ego2Exo and Exo2Ego tasks into a single cross-view mapping framework, enabling the model to jointly learn from both directional data and perform bidirectional correspondence within a unified formulation. This design also enables the application of $\mathcal{L}_{cycle}$.
Second, we introduce same-view exemplar synthesis by asynchronously sampling Ego2Ego and Exo2Exo pairs, which increases data diversity and strengthens intra-view consistency.
Third, for cross-view pairs, we relax temporal alignment by pairing query frames with temporally offset target frames, improving the model’s resilience to timing discrepancies across views.
% To enrich our training data and simplify model design, we employ three complementary processing strategies, each of which is applied jointly during training:

% \begin{itemize}
%     \item \textbf{Unified cross-view training.} We collapse the previously separate \textbf{Ego2Exo} and \textbf{Exo2Ego} tasks into a single formulation by treating both ego-centric$\rightarrow$exo-centric and exo-centric$\rightarrow$ego-centric pairs as instances of a unified cross-view mapping problem. A single network is trained to handle both directions, avoiding the need for separate models.
    
%     \item \textbf{Same-view exemplar synthesis.} To increase sample diversity, we generate \textbf{Ego2Ego} and \textbf{Exo2Exo} pairs, in which the query and target originate from the same viewpoint. These same-view pairs are sampled asynchronously (i.e., without enforcing temporal synchronization), encouraging the model to learn robust intra-view correspondences under varying temporal offsets.
    
%     \item \textbf{Relaxed temporal alignment (cross-view only).} For cross-view pairs (Ego2Exo and Exo2Ego), we relax the original dataset’s strict temporal synchronization by pairing each query–mask pair with an asynchronous target frame. This excludes Ego2Ego and Exo2Exo samples, alleviates over-reliance on exact frame alignment, and promotes temporal robustness across viewpoints.
% \end{itemize}

\paragraph{Evaluation Metrics.}
We adopt the following evaluation metrics following the Ego-Exo4D correspondence benchmark: 1) \textit{Visibility Accuracy (VA)}~\cite{brodersen2010balanced}: Evaluates the model’s ability to predict object visibility in the target view, considering occlusions and out-of-frame cases. The CLS head is specifically designed to optimize this metric. 2) \textit{Intersection over Union (IoU):} Measures the overlap between the predicted and ground-truth masks. 3) \textit{Location Error (LE):} Defined as the normalized distance between the centroids of the predicted and ground-truth masks. 4) \textit{Contour Accuracy (CA)}~\cite{perazzi2016benchmark}: Evaluates mask shape similarity after aligning the centroids of the predicted and ground-truth masks.

According to the evaluation criteria of Ego-Exo4D Correspondence Challenge, the primary evaluation metric is the mean Intersection-over-Union of both Ego2Exo and Exo2Ego task settings. We refer to this metric as \textbf{mIoU}.

\begin{table*}[t]
\centering % 推荐使用 \centering 代替 center 环境用于表格
\addtolength{\tabcolsep}{1pt} % 保留您原来的列间距调整
\vspace{-0.05in}
\caption{\textbf{Evaluation results on the Ego-Exo4D correspondence benchmark v2 (test set).} The best performance is highlighted in \textbf{bold}, and the second-best is \underline{underlined}. %$^\dagger$ Models without training on the Ego-Exo4D correspondence benchmark. 
}
\vspace{-0.05in}
\scalebox{1}{
\begin{tabular}{l cccc cccc c} % 不使用竖线，列格式定义中的 | 被移除
    \toprule % booktabs 的顶部线
    & \multicolumn{4}{c}{\textbf{Ego Query}} & \multicolumn{4}{c}{\textbf{Exo Query}} &  \\
    \cmidrule(lr){2-5} \cmidrule(lr){6-9} % booktabs 的部分横线，(lr) 表示左右留空
    \textbf{Method}  & \textbf{VA}$\uparrow$ & \textbf{IoU}$\uparrow$ & \textbf{LE}$\downarrow$ & \textbf{CA}$\uparrow$
                      & \textbf{VA}$\uparrow$ & \textbf{IoU}$\uparrow$ & \textbf{LE}$\downarrow$ & \textbf{CA}$\uparrow$
                      & \textbf{mIoU}$\uparrow$\\
    \midrule % booktabs 的中间线，分隔表头和数据

    % 修改后的分组标题样式
    %\addlinespace[0.5em] % 在分组标题前增加一些垂直间距
    % \multicolumn{10}{l}{\textbf{Access to one frame per view}} \\ 
    %\addlinespace[0.3em] % 在分组标题后，数据行前增加少量间距

    XSegTx (random weights)~\cite{grauman2024ego}    & 50.00 & 0.48 & 0.118 & 0.014 & 50.00 & 1.08 & 0.203 & 0.024 & 0.78\\
    XSegTx~\cite{grauman2024ego}    & 66.31 & 18.99 & 0.070 & 0.386 & 82.01 & 27.14 & 0.104 & 0.358 & 23.07\\
    XMem~\cite{grauman2024ego,cheng2022xmem}    & 64.39 & 19.28 & 0.151 & 0.262 & 60.35 & 16.56 & 0.160 & 0.204 & 17.92\\
    XView-XMem~\cite{grauman2024ego}    & 61.24 & 14.84 & 0.115 & 0.242 & 61.72 & 21.37 & 0.139 & 0.269 & 18.11\\
    XView-XMem (+ XSegTx)~\cite{grauman2024ego}  & 66.79 & 34.90 & 0.038 & 0.559 & 59.71 & 25.00 & 0.117 & 0.327 & 29.95\\
    SEEM~\cite{zou2023segment} & $\emptyset$ & 1.53 & 0.258 & 0.041 & $\emptyset$ & 4.34 & 0.289 & 0.062 & 2.94\\
    PSALM~\cite{zhang2024psalm} & $\emptyset$ & 7.40 & 0.266 & 0.121 & $\emptyset$ & 2.10 & 0.294 & 0.058 & 4.75\\
    CMX~\cite{zhang2023cmx} & $\emptyset$ & 6.80 & 0.110 & 0.137 & $\emptyset$ & 12.00 & 0.166 & 0.177 & 9.40\\
    ObjectRelator~\cite{fu2024objectrelator} & \underline{95.95} & 35.27 & \underline{0.036} & 0.540 & \underline{97.36} & 40.31 & \textbf{0.068} & 0.500 & 37.79\\
    O-MaMa ~\cite{mur2025mama} & 50.00 & \textbf{42.57} & \textbf{0.033} & \underline{0.590} & 50.05 & \underline{44.08} & 0.082 & \underline{0.524} & \underline{43.32}\\
    \rowcolor{cyan!10}
    Ours & \textbf{98.92} & \underline{41.95} & 0.038 & \textbf{0.669} & \textbf{99.86} & \textbf{47.18} & \underline{0.081} & \textbf{0.591} & \textbf{44.57}\\

    % \addlinespace[0.3em] % 在两组数据之间增加更明显的间距，也可以用 \midrule
    % \hline
    %  \addlinespace[0.3em] % 在分组标题后，数据行前增加少量间距
    % \multicolumn{10}{l}{\textbf{Access to multiple frames per view}} \\ 
    
    \bottomrule % booktabs 的底部线
    %\addlinespace[0.5em]
\end{tabular}
}
\vspace{-0.05in}

\label{tab:main}
\end{table*}

\subsection{Comparison to competitive approaches}

\paragraph{Baselines.}

%To the best of our knowledge, this work presents the first open-source and reproducible approach to the ego-exo correspondence task. As such, the only directly comparable baselines are those introduced by Ego-Exo4D. 
We evaluate against the following 7 open-source models: 1) \textit{XSegTx}: A Transformer-based spatial model adapted from SegSwap~\cite{shen2022learning} that independently estimates correspondences at each time step. 2) \textit{XView-XMem}: A spatio-temporal model adapted from XMem~\cite{cheng2022xmem} that leverages temporal context to generalize object tracking across views using ground-truth masks from one view per frame. 3) \textit{SEEM}: A universal segmentation framework that interprets visual prompts in reference to the input image. 4) \textit{PSALM}: A model extends Phi-1.5 1.3B model~\cite{li2023textbooks} with a mask decoder and a flexible prompting schema to perform diverse pixel-wise segmentation tasks within a single unified framework. 5) \textit{CMX}: A transformer-based segmentation model that integrates information from two modalities. 6) \textit{ObjectRelator}: A cross-view correspondence model that fuses visual and textual cues and aligns representations between egocentric and exocentric views with specialized cross-view modules. 7) \textit{O-MaMa}:  An object mask matching approach that selects the best mask candidate from FastSAM~\cite{zhao2023fast} proposals between images.

\paragraph{Results on Ego-Exo4D.}

The quantitative results on the Ego-Exo4D correspondence benchmark are summarized in Table~\ref{tab:main}.
Despite limited training data, it achieves an mIoU of 44.57\%, representing a +2.9\% relative improvement over the previous SOTA method, O-MaMa.
Our method further surpasses all prior baselines in the Exo Query setup by a +7.0\% relative improvement in IoU and attains competitive performance in the Ego Query setup. The improvement is also consistent across other metrics, with a relative gain of up to +13.4\% in CA under the Ego Query setting. Overall, these results highlight the strong effectiveness and robustness of our approach.

%The quantitative results on the Ego-Exo4D correspondence benchmark are summarized in Table~\ref{tab:main}. Despite using only a single frame per view and less training data, our method achieves performance comparable to the strongest baseline in the Ego Query setting. Notably, our method outperforms this baseline in the Exo Query setting. With respect to the primary evaluation metric, mIoU, our approach improves upon the strongest baseline from 29.95 to \todo{}, highlighting its effectiveness.

%We observe that our method and XSegTx perform relatively better in the Exo Query setting. A plausible explanation is that ground-truth masks in the Ego view are generally larger, potentially making training and prediction easier. In contrast, models based on XMem show stronger performance in the Ego view, likely due to the more significant view changes in this setting. These changes may reduce the utility of the additional temporal information provided by multiple frames, which appears more beneficial in the Exo view without view change.

We observe that most methods perform worse under the Ego Query setting. This observation is consistent with intuition, as exo-view objects are generally smaller and appear within more cluttered environments, making segmentation more challenging. The judgement is quantitatively validated in Section~\ref{sec:ablation} and Figure~\ref{fig:task_size_compare} (b). Meanwhile, although SEEM and PSALM demonstrate strong performance on other out-of-domain tasks, they struggle to generalize effectively in this setting. This suggests that training on cross-view datasets is crucial for achieving robust performance, as knowledge learned from other domains does not readily transfer to cross-view correspondence.

% Interestingly, we observe that our method and XSegTx tend to perform better in the Exo Query setting, possibly because object masks in the Ego view are generally larger, making training and prediction relatively easier. Conversely, XMem-based models benefit more from the Ego view, likely due to their ability to exploit temporal information—less disrupted by viewpoint shifts—which is more stable in the Ego perspective than the Exo. Our results suggest that strong spatial modeling, even with minimal input, can be highly effective across both views.

\paragraph{Results on HANDAL-X.} 
Since ObjectRelator additionally reports results on HANDAL-X, we also include evaluations on this dataset without test-time training. As shown in Table~\ref{tab:handal}, our approach exhibits superior cross-view generalization, achieving an IoU of 78.8\% on HANDAL-X without any training on this dataset, surpassing all baselines by a relative improvement of 84.1\%. After fine-tuning on HANDAL-X, our method continues to consistently outperform all competing approaches. These results can be attributed to our compact design, along with the effective training data augmentation strategy.

Notably, all methods exhibit performance gains on HANDAL-X after being trained on Ego-Exo4D, indicating that exposure to egocentric–exocentric data effectively enhances generalization to unseen cross-view configurations, even under distinct scenes and viewpoints.

\begin{table}[!t]
    \centering
    \caption{{\bf Evaluation results on the HANDAL-X benchmark.} }
    \vspace{-0.05in}
    \scalebox{1}{
    \begin{tabular}{l c c}
    \toprule
    \textbf{Method} &  \textbf{Fine-tuning Datasets} & \textbf{IoU}$\uparrow$  \\ 
    \midrule
    XSegTx~\cite{grauman2024ego} & $\emptyset$ & 1.5 \\
    SEEM~\cite{zou2023segment} &  $\emptyset$ & 2.5 \\
    PSALM~\cite{zhang2024psalm} & $\emptyset$ & 14.2 \\
    PSALM~\cite{zhang2024psalm} & Ego-Exo4D & 39.9 \\
    ObjectRelator~\cite{fu2024objectrelator} & Ego-Exo4D & 42.8 \\
    \cellcolor{cyan!10}Ours & \cellcolor{cyan!10}Ego-Exo4D & \cellcolor{cyan!10}\textbf{78.8} \\
    \midrule
    PSALM~\cite{zhang2024psalm} & Ego-Exo4D, HANDAL-X & 83.4 \\
    ObjectRelator~\cite{fu2024objectrelator} & Ego-Exo4D, HANDAL-X & 84.7 \\
    \cellcolor{cyan!10}Ours & \cellcolor{cyan!10}Ego-Exo4D, HANDAL-X & \cellcolor{cyan!10}\textbf{85.0} \\
    \bottomrule
    \end{tabular}
    }
    \vspace{-0.05in}
    
    \label{tab:handal}
\end{table}

\subsection{Ablation Study}
\label{sec:ablation}
% \paragraph{Loss Function Design.} As is mentioned in Section \ref{subsec:loss}, we introduce two kinds of losses additional to the loss design of XSegTx: Cycle Consistency Loss and Auxiliary Loss. Our results show that none of the newly introduced losses contributes anything to our method.
% \begin{table}[h]
% \begin{center}
% \begin{tabular}{l | c c c | c c c}
% \textbf{Loss Design} & $\mathcal{L}_{cycle}$ & $\mathcal{L}_{aux}$ & TTT & \textbf{Ego-IoU}$\uparrow$ & \textbf{Exo-IoU}$\uparrow$ & \textbf{mIoU}$\uparrow$ \\
% \midrule
% \rowcolor{LightGray}
% Ours                        & \cmark & \cmark & \cmark & \textbf{32.39} & -- & -- \\
% Ours w.o TTT        & \cmark & \cmark & \xmark & 31.87 & 39.82 & 35.85 \\
% Ours w.o TTT \& $\mathcal{L}_{cycle}$ & \xmark & \cmark & \xmark & 31.42 & 39.74 & 35.58 \\
% Ours w.o TTT \& $\mathcal{L}_{aux}$   & \cmark & \xmark & \xmark & -- & -- & -- \\
% \end{tabular}
% \end{center}
% \caption{\textbf{Ablation Study on Optimization Components.} }
% \label{tab:abl_loss}
% \end{table}

\begin{figure}
\centering
\includegraphics[width=0.48\textwidth]{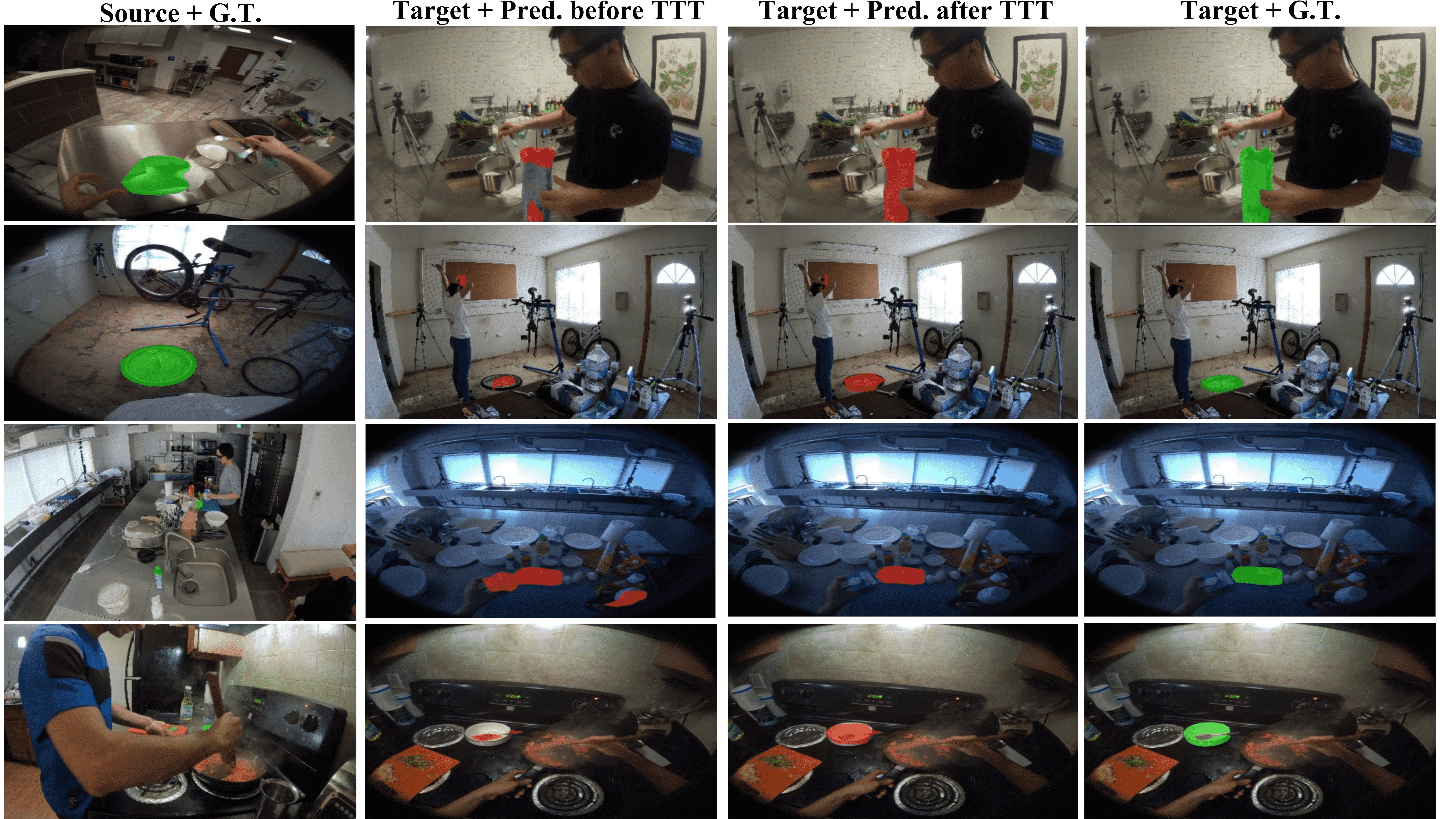}
\vspace{-0.05in}
\caption{
{\bf Visualization illustrating the contribution of test-time training.}
}
\vspace{-0.05in}
\label{fig:qualitative_ttt}
\end{figure}

\paragraph{Loss Components.} As described in Section~\ref{subsec:loss}, we enhance the baseline loss $\mathcal{L}_{\text{mask}}$ with a cycle consistency loss $\mathcal{L}_{\text{cycle}}$ and an auxiliary loss $\mathcal{L}_{\text{aux}}$. As shown in Table~\ref{tab:abl_loss}, removing either component causes a performance degradation, confirming their effectiveness. Test-time tuning (TTT) further brings consistent improvements. Notably, removing $\mathcal{L}_{\text{cycle}}$ leads to a noticeable drop in mIoU, highlighting its importance in providing the self-supervised signal essential for effective TTT. As illustrated in Figure~\ref{fig:qualitative_ttt}, after applying TTT, the model not only focuses more accurately on the true target object while suppressing visually similar distractors but also produces predicted masks that more completely cover the ground-truth regions.

\begin{table}[t]
\begin{center}
\caption{\textbf{Ablation on optimization components.} }
\label{tab:abl_loss}
\vspace{-0.05in}
\resizebox{1\linewidth}{!}{
\begin{tabular}{l c c c c c c}
\toprule
\textbf{Configuration} & $\mathcal{L}_{\text{cycle}}$ & $\mathcal{L}_{\text{aux}}$ & TTT & \textbf{Ego-IoU}$\uparrow$ & \textbf{Exo-IoU}$\uparrow$ & \textbf{mIoU}$\uparrow$ \\
\midrule
\rowcolor{cyan!10}
Ours (full) & \cmark & \cmark & \cmark & \textbf{41.95} & \textbf{47.18} & \textbf{44.57} \\
w/o $\mathcal{L}_{\text{cycle}}$ & \xmark & \cmark & \cmark & 40.28 & 45.82 & 43.05 \\
w/o $\mathcal{L}_{\text{aux}}$ & \cmark & \xmark & \cmark & 40.64 & 43.81 & 42.90 \\
w/o TTT & \cmark & \cmark & \xmark & 41.79 & 44.18 & 42.99 \\
% w/o TTT \& $\mathcal{L}_{cycle}$ & \xmark & \cmark & \xmark & 31.42 & 39.74 & 35.58 \\
% w/o TTT \& $\mathcal{L}_{aux}$ & \cmark & \xmark & \xmark & 31.68 & 39.34 & 35.51 \\
% w/o $\mathcal{L}_{cycle}$ \& $\mathcal{L}_{aux}$ & \xmark & \xmark & \cmark & 31.68 & 39.34 & 35.51 \\
% w/o TTT \& $\mathcal{L}_{aux}$ \& $\mathcal{L}_{cycle}$ & \xmark & \xmark & \xmark & 31.79 & 39.51 & 35.65 \\
\bottomrule
\end{tabular}
}
\end{center}
\vspace{-0.05in}
\end{table}

% \begin{table}[h]
% \begin{center}
% \begin{tabular}{c c c | c}
% \textbf{\( lr_{\text{ttt}} \)} & \textbf{\( T \)} & \textbf{\( K \)} & \textbf{mIoU}$\uparrow$ \\
% \midrule
% \rowcolor{LightGray}
%         $5\times 10^{-6}$ & 3 & 3 & \textbf{34.46} \\ 
% 	$1\times 10^{-5}$ & 3 & 3 & 34.03 \\ 
%         $1\times 10^{-6}$ & 3 & 3 & 34.21 \\
% 	$5\times 10^{-6}$ & 4 & 3 & 34.40 \\
%         $5\times 10^{-6}$ & 2 & 3 & 34.32 \\ 
%         $5\times 10^{-6}$ & 3 & 4 & 34.39 \\
%         $5\times 10^{-6}$ & 3 & 2 & 34.32 \\
        
% \end{tabular}
% \end{center}
% \caption{\textbf{Ablation of Hyperparameters of Test-Time Training.} }
% \label{tab:abl_ttt}
% \end{table}

% \paragraph{Unified cross-view training.} All previously reported results are based on training the Ego2Exo and Exo2Ego models independently, each using its corresponding training split. To this end, 

% \begin{table}[h]
% \begin{center}
% \begin{tabular}{c | c c}
% 	\textbf{Training data} & \textbf{Ego-IoU}$\uparrow$ & \textbf{Exo-IoU}$\uparrow$ \\
% 	\midrule
% 	\rowcolor{LightGray}
% 	Ours (Ego2Exo + Exo2Ego) & \textbf{31.87} & \textbf{39.82} \\ 
% 	\hspace{3mm} Ego2Exo & 31.51 & --\\ 
% 	\hspace{3mm} Exo2Ego & -- & 39.79\\ 
% \end{tabular}
% \end{center}
% \caption{\textbf{Results of Unified Cross-view Training. }}
% \label{tab:abl_data}
% \end{table}

\paragraph{Data Augmentation.} To quantify the individual contributions of each data preprocessing strategy, we conduct a comprehensive ablation study. Since ObjectRelator has already validated the effectiveness of jointly learning from both Ego2Exo and Exo2Ego tasks, we omit this experiment. As shown in Table~\ref{tab:abl_data}, removing any single strategy consistently degrades both per-view IoU and overall mIoU, confirming that each component contributes to more robust and generalizable segmentation.

\begin{table}
\begin{center}
\caption{\textbf{Ablation on data augmentation. } "RTA" refers to the relaxed temporal alignment strategy used for both tasks.}
\vspace{-0.05in}
\label{tab:abl_data}
\resizebox{1\linewidth}{!}{
\begin{tabular}{l c c c}
    \toprule
	\textbf{Training data} & \textbf{Ego-IoU}$\uparrow$ & \textbf{Exo-IoU}$\uparrow$ & \textbf{mIoU}$\uparrow$ \\
	\midrule
	\rowcolor{cyan!10}
	Ours  & \textbf{41.95} & \textbf{47.18} & \textbf{44.57}\\ 
    \hspace{3mm} w/o Ego2Ego \& Exo2Exo & 40.88 & 45.50 & 43.19\\ 
	\hspace{3mm} w/o RTA & 40.60 & 45.45 & 43.03 \\ 
    \bottomrule
\end{tabular}
}
\end{center}
\vspace{-0.05in}
\end{table}

\paragraph{Contribution beyond DINOv3 features.}
To isolate and highlight the contribution of our framework beyond the use of DINOv3 features, we adopt XSegTx~\cite{grauman2024ego} from Ego-Exo4D as a baseline and replace its original backbone with DINOv3 for controlled comparison. The corresponding results are reported in Table~\ref{tab:design_advantage}. Notably, even with slightly weaker DINOv2 features, our method still outperforms “baseline + DINOv3,” indicating that the improvements stem primarily from our architectural components rather than the choice of pre-trained features.

\begin{table}[t]
\begin{center}
\vspace{-0.05in}
\caption{\textbf{Disentangling feature quality from method design.} CBS denotes the proposed conditional binary segmentation.}
\label{tab:design_advantage}
\vspace{-0.05in}
\resizebox{1\linewidth}{!}{
\begin{tabular}{l l c c c c c c}
\toprule
\textbf{Framework} & \textbf{Backbone} & \textbf{CBS} & $\mathcal{L}_{\text{cycle}}$ & TTT & \textbf{Ego-IoU}$\uparrow$ & \textbf{Exo-IoU}$\uparrow$ & \textbf{mIoU}$\uparrow$ \\
\midrule
%\rowcolor{cyan!10}
XSegTx & DINOv3 & \xmark & \xmark & \xmark & 26.52 & 34.36 & 30.44 \\
\rowcolor{cyan!10}
Ours & DINOv3 & \cmark & \cmark & \cmark & \textbf{41.95} & \textbf{47.18} & \textbf{44.57} \\
Ours & DINOv2 & \cmark & \cmark & \cmark & 41.48 & 44.50 & 42.99 \\
\bottomrule
\end{tabular}
}
\end{center}
\vspace{-0.05in}
\end{table}

\paragraph{Dice Loss in $\mathcal{L}_{\text{cycle}}$ and TTT.}
It is intuitive to incorporate Dice loss into the self-supervised learning objective, i.e., setting $\mathcal{L}_{\text{cycle}} = \mathcal{L}_{\text{bce}}(M_s, \hat{M}_s) + \lambda_{\text{dice}}^{\prime} \mathcal{L}_{\text{dice}}(M_s, \hat{M}_s)(\lambda_{\text{dice}}^{\prime}=5)$. To maintain a balanced optimization, we additionally explore a configuration with $\lambda_{\text{cycle}}=1$. As reported in Table~\ref{tab:abl_dice_ttt}, the results indicate that introducing Dice loss into the optimization objective hinders the model’s ability to learn effectively during TTT.

\begin{table}[h]
\begin{center}
\vspace{-0.05in}
\caption{\textbf{Ablation on dice loss in $\mathcal{L}_{\text{cycle}}$ and TTT.} }
\label{tab:abl_dice_ttt}
\vspace{-0.05in}
\begin{tabular}{c c c c c}
\toprule
\textbf{$\lambda_{\text{dice}}^{\prime}$} &  \textbf{$\lambda_{\text{cycle}}$} & \textbf{Ego-IoU}$\uparrow$  & \textbf{Exo-IoU}$\uparrow$ & \textbf{mIoU}$\uparrow$ \\
\midrule
\rowcolor{cyan!10}
0 & 10 & \textbf{41.95} & \textbf{47.18} & \textbf{44.57} \\
5 & 10 & 38.25 & 42.14 & 40.20 \\
5 & 1 & 40.45 & 44.57 & 42.51 \\
\bottomrule
\end{tabular}
\end{center}
\vspace{-0.05in}
\end{table}

\paragraph{Linear Probing Stage.}  We conduct an ablation on the first training stage, as shown in Table~\ref{tab:abl_train}. The results demonstrate that this stage is essential for achieving accurate object segmentation, as it stabilizes cross-view feature alignment and provides a strong initialization for subsequent end-to-end fine-tuning.

\begin{table}[h]
\begin{center}
\vspace{-0.05in}
\caption{\textbf{Ablation on the linear probing stage.} }
\label{tab:abl_train}
\vspace{-0.05in}
\resizebox{1\linewidth}{!}{
\begin{tabular}{l c c c}
\toprule
\textbf{Setting} & \textbf{Ego-IoU}$\uparrow$  & \textbf{Exo-IoU}$\uparrow$ & \textbf{mIoU}$\uparrow$ \\
\midrule
\rowcolor{cyan!10}
Ours & \textbf{41.95} & \textbf{47.18} & \textbf{44.57} \\
\hspace{3mm} w/o Linear Probing Stage & 37.97 & 43.69 & 40.83 \\
\bottomrule
\end{tabular}
}
\end{center}
\vspace{-0.05in}
\end{table}

\begin{figure}
\centering
\includegraphics[width=0.48\textwidth]{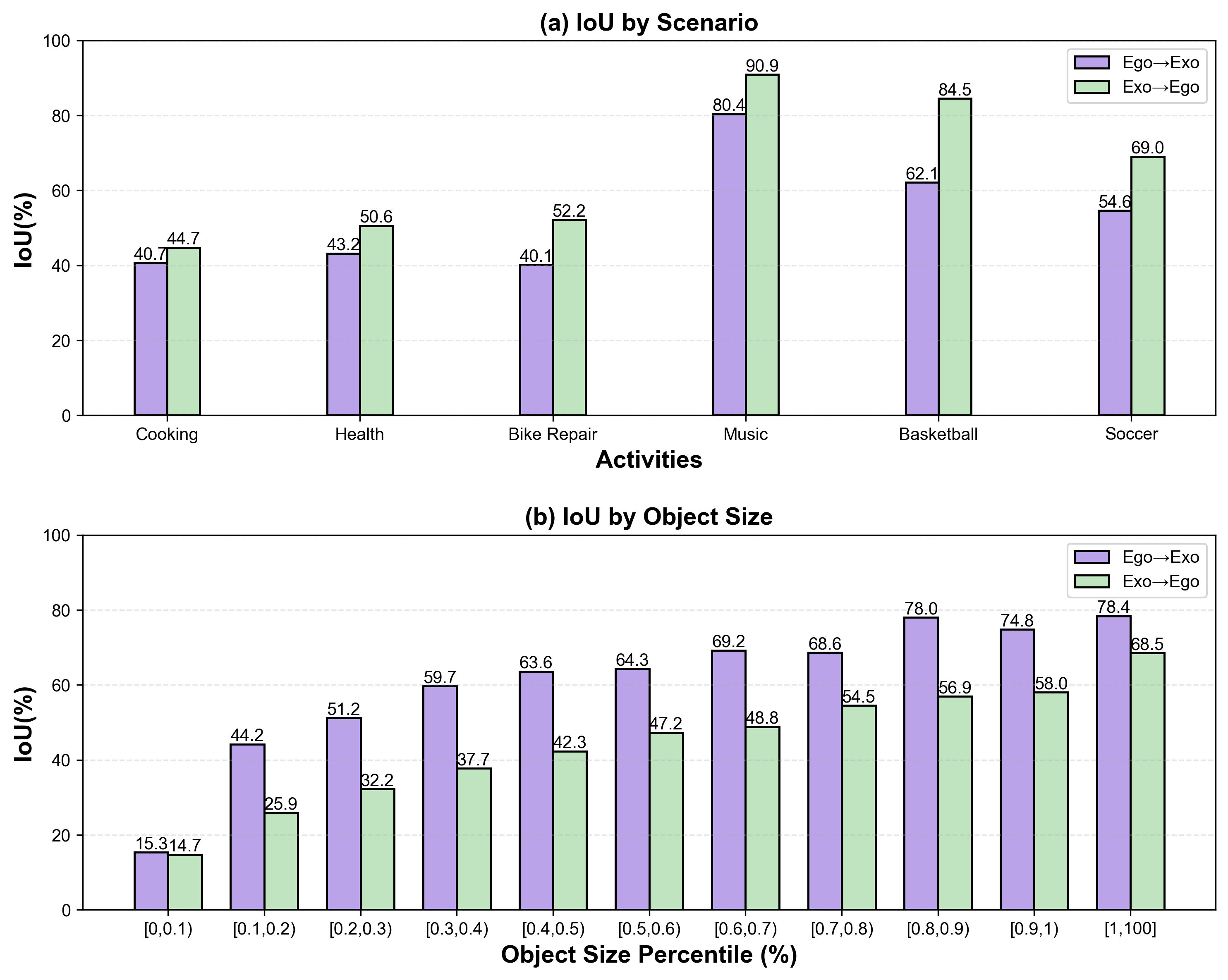}
\vspace{-0.15in}
\caption{{\bf (a) Performance per activity scenario; (b) Performance across different object sizes in the target view. } 
}
\vspace{-0.05in}
\label{fig:task_size_compare}
\end{figure}

\paragraph{Results across Different Scenarios.}
The test set of the Ego-Exo4D Correspondence Benchmark includes 6 distinct scenarios. As shown in Figure~\ref{fig:task_size_compare} (a), cooking, health, and bike repair present greater challenges due to smaller object sizes and more complex environments. Nevertheless, our method consistently achieves an IoU exceeding 40\% across all scenarios, highlighting its robustness in handling diverse objects and environments.

\paragraph{Results across Different Target Object Sizes.} 
We evaluate our method across different object sizes in the target view, as shown in Figure~\ref{fig:task_size_compare} (b). Our method performs well on objects occupying more than 0.1\% of the target-view image area, while smaller objects remain challenging.

It is worth noting that although the IoU of the Ego2Exo task is consistently higher than that of the Exo2Ego task within each object-size bin, the overall IoU of Ego2Exo (41.95\%) remains lower than that of Exo2Ego (47.18\%). This discrepancy suggests that the two tasks differ in their underlying object-size distributions. Ego2Exo likely contains a larger proportion of smaller target objects, which causes most methods including ours to yield lower IoU and also limits the effectiveness of TTT, as reflected in Table~\ref{tab:main} and Table~\ref{tab:abl_loss}.

% \begin{figure}
% \centering
% \includegraphics[width=0.48\textwidth]{fig/size_compare.png}
% \vspace{-0.05in}
% \caption{
% {\bf Performance across Different Object Sizes in the Target View.} 
% }
% \vspace{-0.05in}
% \label{fig:size_compare}
% \end{figure}

\subsection{Qualitative Results}

\begin{figure}
\centering
\includegraphics[width=0.48\textwidth]{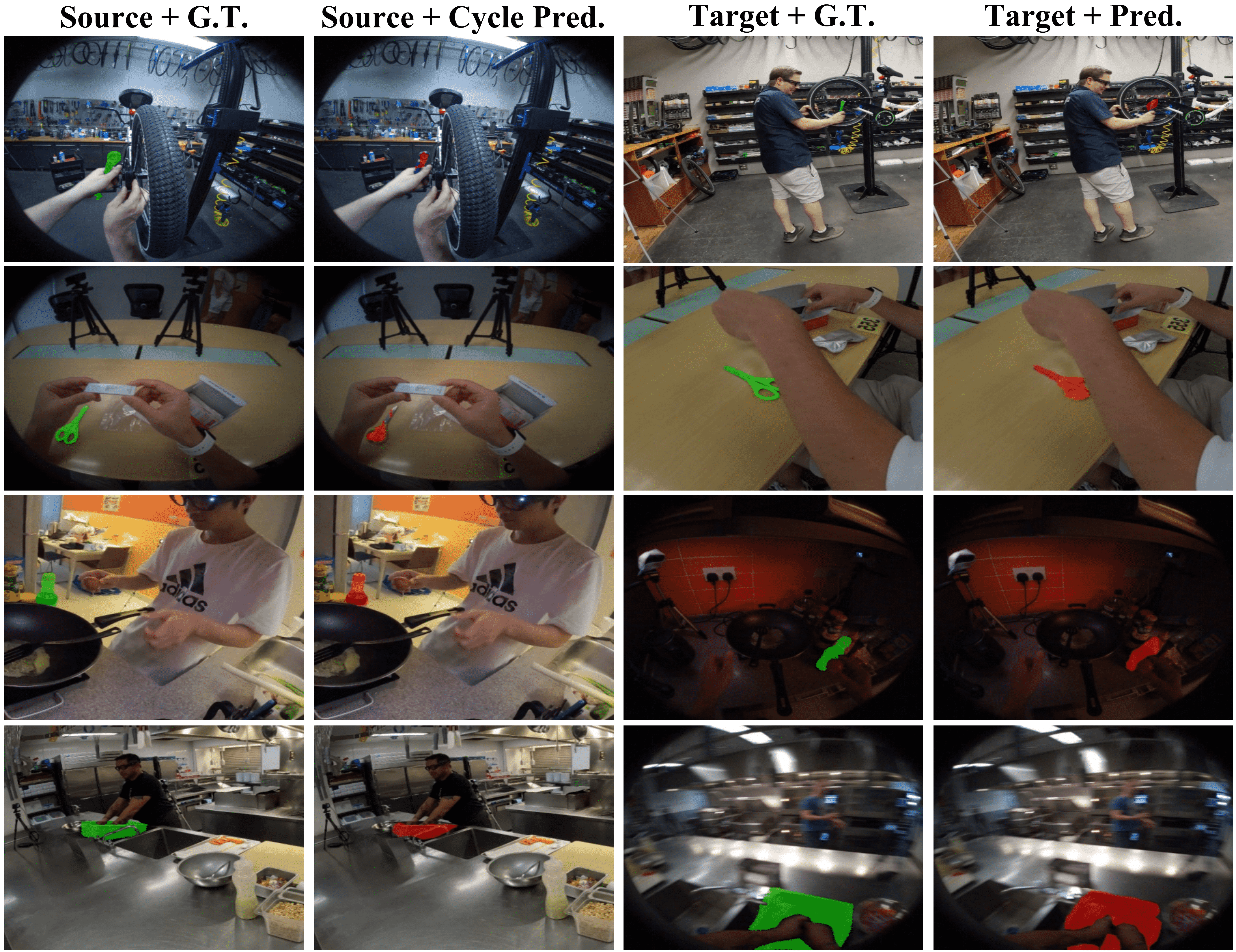}
\vspace{-0.05in}
\caption{
{\bf Qualitative results on the Ego-Exo4D correspondence benchmark.}
Each row corresponds to one sample. From top to bottom, the first and second rows show samples of Ego2Exo, while the third and fourth rows show samples of Exo2Ego. 
}
\vspace{-0.05in}
\label{fig:qualitative}
\end{figure}

\begin{figure}
\centering
\includegraphics[width=0.48\textwidth]{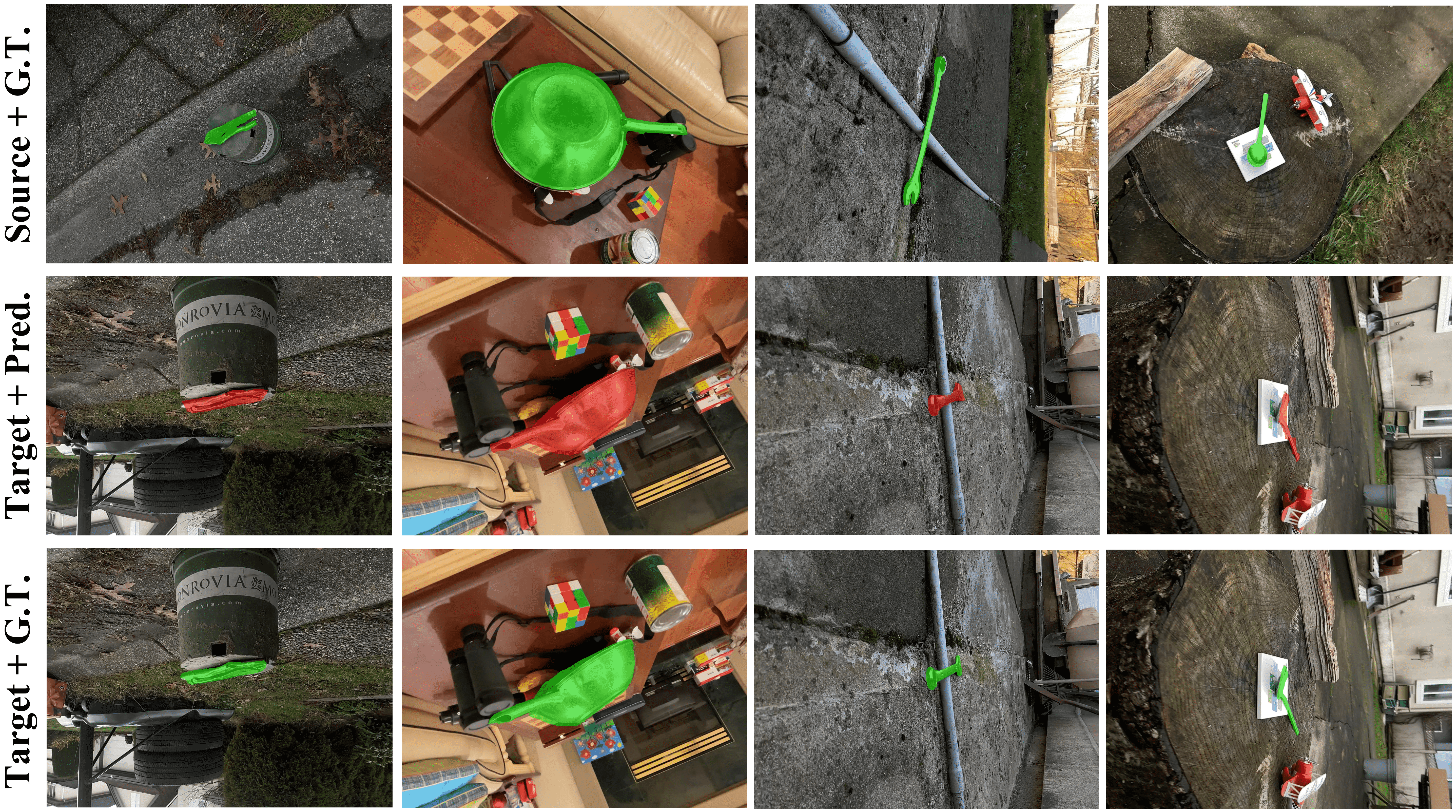}
\vspace{-0.05in}
\caption{
{\bf Qualitative results on the HANDAL-X benchmark.} Each column corresponds to one sample.
}
\vspace{-0.05in}
\label{fig:qualitative_handal}
\end{figure}

To better demonstrate the effectiveness of our pipeline, we present representative qualitative results on Ego-Exo4D in Figure~\ref{fig:qualitative}, covering both Ego2Exo and Exo2Ego scenarios, and on HANDAL-X in Figure~\ref{fig:qualitative_handal}. These visual examples highlight the challenging nature of the task, with large variations in scale, perspective, and object appearance due to deformation or occlusion. Despite these difficulties, our method consistently and accurately segments the queried object across diverse activities and viewpoints. The results demonstrate not only the robustness of our approach in handling extreme viewpoint shifts and motion but also support the effectiveness of using a cycle-consistent loss, which encourages reliable cross-view alignment by enforcing mutual consistency between Ego and Exo predictions. More visual examples and failure cases are provided in the supplementary material.

% \paragraph{Limitation.} While test-time tuning (TTT) consistently improves performance, it introduces additional computational overhead during inference. In our current (unoptimized) implementation, inference with TTT takes longer compared to the vanilla setting (360 s vs. 30 s per take). We believe this cost can be significantly reduced through further engineering optimizations or by adopting more lightweight tuning strategies.
\section{Conclusion}

We present a simple yet effective approach for object correspondence between egocentric and exocentric views. By unifying cross-view tasks, enriching training with same-view and temporally relaxed pairs, and leveraging a cycle-consistent loss with test-time training, our method achieves strong performance on the Ego-Exo4D and HANDAL-X benchmarks. Extensive experiments demonstrate its robustness to viewpoint shift, occlusion, and motion. Our findings suggest that careful training design and self-supervision can enhance correspondence without requiring complex pipelines or large temporal context. We hope our work offers meaningful insights and a solid foundation that facilitates future research on this \text{underexplored task}.

\section*{Acknowledgment}

 This work was supported by the National Key R\&D Program of China (2022YFB4701400/4701402), SSTIC Grant(KJZD20230923115106012,  KJZD20230923114916032, GJHZ20240218113604008).

{
    \small
    \bibliographystyle{ieeenat_fullname}
    \bibliography{main}
}

\clearpage
\setcounter{page}{1}
\maketitlesupplementary
\appendix

\section{Organization} \label{sec:org}

This document contains the following sections: 
\begin{itemize}[leftmargin=2em]
	\item More training details are provided in Section~\ref{sec:train_detail}.  
	
	%\item Details of visibility prediction are provided in Section~\ref{sec:va_detail}.

    \item Test-time training on HANDAL-X is provided in Section~\ref{sec:ttt_handalx}.

    \item More ablation study is provided in Section~\ref{sec:ablation}.

    \item Efficiency analysis is provided in Section~\ref{sec:efficiency}

    \item More qualitative results are provided in Section~\ref{sec:more_vis}.

    \item Limitation and future work are provided in Section~\ref{sec:limit}.
	
\end{itemize}
All blue-highlighted rows in the tables denote the default configurations of our method. We refer to test-time training as \textbf{TTT} throughout the paper.

\section{More Training Details}
\label{sec:train_detail}

We train our model using the AdamW optimizer~\cite{loshchilovdecoupled} with a cosine learning rate schedule and linear warm-up. We use a batch size of 16. The image size is $512\times512$. 
The training process on Ego-Exo4D~\cite{grauman2024ego} consists of two stages. In the first stage (\textit{linear probing}), we freeze the two DINOv3~\cite{simeoni2025dinov3} backbones and train the remaining modules for 64K iterations. The learning rate decays from the maximum value of $1\times10^{-3}$ to a minimum of $1\times10^{-4}$. 
In the second stage, all parameters are unfrozen and optimized for 640K iterations. The learning rate decays from the maximum value of $1\times10^{-5}$ to a minimum of $1\times10^{-6}$. 
To address GPU memory limitations (40GB), we adopt gradient accumulation with a step size of 16, resulting in an effective number of parameter updates of 704K / 16 = 44K. The training process takes approximately 72 hours on 8 NVIDIA RTX A800 GPUs. We maintain an exponential moving average (EMA) of the model parameters throughout training, and use the EMA model as the final model for evaluation. 
For visibility prediction, we fine-tune only the CLS Head for 96K iterations, which takes approximately 1 hour on the same hardware, using the same training setup as the main binary segmentation task. 
In the TTT stage, the adaptation takes approximately 3 hours for Ego2Exo and 12 hours for Exo2Ego.

On HANDAL-X, we train for 10 epochs with the learning rate decaying from the maximum value of $2\times10^{-4}$ to a minimum of $2\times10^{-6}$. The offline training stage requires approximately 2 hours, and the TTT stage requires an additional 1 hour.

% \section{Details of Visibility Prediction}
% \label{sec:va_detail}

% Without the CLS Head and CLS token, a natural way to infer visibility is through the dense mask output, i.e., if the predicted mask is entirely zero, the object is considered invisible. However, in practice, we found that relying on dense predictions for this coarse-level task is highly unstable and leads to significant errors.

% It is notable that most methods in Table 1 of the main text achieve only 50–69 visibility accuracy (VA)~\cite{brodersen2010balanced}, measured using balanced accuracy. This is largely due to the extreme scarcity of invisible-object samples in the training set, making it difficult for existing approaches to reliably determine whether the target-view object is invisible. To address this limitation, we introduce a post-trained CLS Head, which naturally aligns with classification tasks and image-level reasoning. During post-training, we fine-tune the CLS Head using a balanced sampling strategy (visible : invisible = 1:1), which substantially enhances its ability to correctly identify invisible targets, as demonstrated in Table~\ref{tab:abl_posttrain}.

% \begin{table}[h]
% \begin{center}
% \caption{\textbf{Ablation on the post-training stage.} }
% \label{tab:abl_posttrain}
% \begin{tabular}{l c c}
% \toprule
% \textbf{Setting} & \textbf{Ego-VA}$\uparrow$  & \textbf{Exo-VA}$\uparrow$ \\
% \midrule
% \rowcolor{cyan!10}
% Ours & \textbf{98.92} & \textbf{99.86} \\
% \hspace{3mm} w/o post-training stage & 50.00 & 50.02 \\
% \bottomrule
% \end{tabular}
% \end{center}
% \end{table}

\section{Test-time Training on HANDAL-X}
\label{sec:ttt_handalx}

Since our results on the HANDAL-X benchmark~\cite{fu2024objectrelator} without TTT already surpass all baselines, we omit the TTT results from the main text. Table~\ref{tab:handal_ttt} presents the quantitative performance with TTT on HANDAL-X, further demonstrating its effectiveness and generalization across benchmarks. We observe that when the baseline IoU is already very high, TTT yields only marginal improvements. The corresponding qualitative results are provided in Figure~\ref{fig:app_qualitative_handalx}.

\begin{table}[!t]
    \centering
    \caption{{\bf Evaluation results on the HANDAL-X benchmark.} }
    \scalebox{1}{
    \begin{tabular}{l c c}
    \toprule
    \textbf{Method} &  \textbf{Fine-tuning Datasets} & \textbf{IoU}$\uparrow$  \\ 
    \midrule
    XSegTx~\cite{grauman2024ego} & $\emptyset$ & 1.5 \\
    SEEM~\cite{zou2023segment} &  $\emptyset$ & 2.5 \\
    PSALM~\cite{zhang2024psalm} & $\emptyset$ & 14.2 \\
    PSALM~\cite{zhang2024psalm} & Ego-Exo4D & 39.9 \\
    ObjectRelator~\cite{fu2024objectrelator} & Ego-Exo4D & 42.8 \\
    Ours (w/o TTT) & Ego-Exo4D & 78.8 \\
    \cellcolor{cyan!10}Ours & \cellcolor{cyan!10}Ego-Exo4D & \cellcolor{cyan!10}\textbf{80.6} \\
    \midrule
    PSALM~\cite{zhang2024psalm} & Ego-Exo4D, HANDAL-X & 83.4 \\
    ObjectRelator~\cite{fu2024objectrelator} & Ego-Exo4D, HANDAL-X & 84.7 \\
    Ours (w/o TTT) & Ego-Exo4D, HANDAL-X & 85.0 \\
    \cellcolor{cyan!10}Ours & \cellcolor{cyan!10}Ego-Exo4D, HANDAL-X & \cellcolor{cyan!10}\textbf{85.3} \\
    \bottomrule
    \end{tabular}
    }    
    \label{tab:handal_ttt}
\end{table}

\section{More Ablation Study}
\label{sec:ablation}

\paragraph{Mask Prediction Method.}

To enable the model to adaptively predict segmentation masks conditioned on given object features, we further explore an alternative implementation named \textit{Cosine Prediction} for mask generation. The final segmentation mask \( \hat{M}_t \) is predicted using both visual tokens and the updated condition token \( y_{\text{cdt}} \). Specifically, for the \( i \)-th visual token \( y_i \), the prediction is computed as:

\begin{equation}
\label{sigmoid_loss}
    \hat{M}^i_t = \mathtt{Sigmoid} \left( \tau \cdot \mathtt{Cos}(y_{\text{cdt}}, y_i) - \beta \right),
\end{equation}
where \( \mathtt{Cos}(\cdot, \cdot) \) denotes cosine similarity, and \( \tau \) and \( \beta \) are learnable temperature and bias parameters as in~\cite{zhai2023sigmoid}. They are initialized to 10 and 5, respectively. 

Table~\ref{tab:abl_pred} presents an ablation study comparing the proposed variant with our original method. %described in Section~\ref{subsec:pipeline}. 
The results demonstrate that direct mask prediction yields better performance than predicting masks conditioned on object features.

\begin{table}[h]
\begin{center}
\caption{\textbf{Ablation on the mask prediction method.} }
\label{tab:abl_pred}
\begin{tabular}{l c c c}
\toprule
\textbf{Method} & \textbf{Ego-IoU}$\uparrow$  & \textbf{Exo-IoU}$\uparrow$ & \textbf{mIoU}$\uparrow$ \\
\midrule
\rowcolor{cyan!10}
Ours & \textbf{41.95} & \textbf{47.18} & \textbf{44.57} \\
Cosine Prediction & 40.29 & 46.75 & 43.52 \\
\bottomrule
\end{tabular}
\end{center}
\end{table}

\paragraph{Dice weight.}
We investigate the influence of the Dice Loss weight \(\lambda_{\text{dice}}\) in our mask supervision objective \(\mathcal{L}_{\text{mask}}\). The Dice Loss \(\mathcal{L}_{\text{dice}}\) plays an essential role, particularly in scenarios where the target occupies only a small region of the spatial mask, as it effectively addresses class imbalance and encourages better alignment of predicted and ground-truth masks. Table~\ref{tab:abl_dice} presents a detailed ablation of performance across different values of \(\lambda_{\text{dice}}\). We find that setting \(\lambda_{\text{dice}} = 5\) yields the best overall performance, outperforming all other configurations. Notably, when \(\lambda_{\text{dice}} = 0\), which effectively removes the Dice Loss, the performance drops significantly across all metrics, confirming the importance of including \(\mathcal{L}_{\text{dice}}\). These results highlight the importance of balancing the Dice component within the mask loss. 

\begin{table}[h]
\begin{center}
\caption{\textbf{Ablation of dice weight.}}
\label{tab:abl_dice}
\begin{tabular}{c c c  c}
\toprule
\textbf{\( \lambda_{\text{dice}} \)} &  \textbf{Ego-IoU}$\uparrow$ & \textbf{Exo-IoU}$\uparrow$ & \textbf{mIoU}$\uparrow$ \\
\midrule
        0 &  28.22 & 32.76 & 30.49\\ 
        0.5 & 41.11 & 46.24 & 43.68 \\
	1 & 41.84 & 47.17 & 44.51 \\ 
        2 & 41.33 & 46.70 & 44.02 \\
        \rowcolor{cyan!10}
        5 & \textbf{41.95} & \textbf{47.18} & \textbf{44.57} \\
	10 & 41.54 & 46.72 & 44.13 \\
\bottomrule
\end{tabular}
\end{center}

\end{table}

\paragraph{Gradian Update Steps of TTT.} 
It is notable that for TTT on Ego-Exo4D, we update for $T{=}2$ steps in Ego2Exo but $T{=}6$ steps in Exo2Ego. To justify this choice, we conduct an ablation study, and the results are presented in Table~\ref{tab:tttsteps_ego} and Table~\ref{tab:tttsteps_exo}. We observe that Ego2Exo achieves its best performance with only 2 update steps, after which further updates cause slight degradation. In contrast, Exo2Ego continues to benefit from additional updates and reaches its peak performance at 7 or more steps. Considering the tradeoff between efficiency and performance, we adopt $T{=}6$ steps as our default setting. These findings highlight the importance of tuning the number of update steps for each direction individually, as the two tasks differ in their underlying object-size distributions and consequently in their adaptation behavior.

\begin{table}[t]
\centering

\begin{minipage}{0.45\linewidth}
\centering
\caption{{\bf Ablation of TTT steps (Ego2Exo).}}
\label{tab:tttsteps_ego}
\begin{tabular}{c c}
\toprule
\textbf{Steps} & \textbf{Ego-IoU}$\uparrow$  \\
\midrule
1  & 41.91 \\
\rowcolor{cyan!10}
2 & \textbf{41.95} \\
3   & 41.90  \\
4   & 41.88 \\
5   &  41.84\\
\bottomrule
\end{tabular}

\end{minipage}
\hfill
\begin{minipage}{0.45\linewidth}
\centering
\caption{{\bf Ablation of TTT steps (Exo2Ego).}}
\label{tab:tttsteps_exo}
\begin{tabular}{c c}
\toprule
\textbf{Steps} & \textbf{Exo-IoU}$\uparrow$  \\
\midrule
3  & 46.81 \\
4 & 46.98 \\
5   & 47.09 \\
\rowcolor{cyan!10}
6  & 47.18 \\
7  & \textbf{47.23} \\
\bottomrule
\end{tabular}
\end{minipage}

\end{table}

\paragraph{Fine-tuning Layes of TTT.}
It is notable that for TTT on Ego-Exo4D, we update the last $K{=}4$ layers in Ego2Exo but $K{=}11$ layers in Exo2Ego. To justify this choice, we conduct an ablation study, and the results are reported in Table~\ref{tab:tttlayers_ego} and Table~\ref{tab:tttlayers_exo}. We observe that Ego2Exo achieves its best performance when only a small number of layers are adapted, while deeper adaptation yields diminishing returns or slight degradation. In contrast, Exo2Ego benefits from updating a substantially larger portion of the network, with performance peaking at 11 layers. These findings suggest that Ego2Exo requires only lightweight adjustments for effective adaptation, whereas Exo2Ego demands broader model capacity to accommodate the larger cross-view domain gap.

\begin{table}[t]
\centering

\begin{minipage}{0.45\linewidth}
\centering
\caption{{\bf Ablation of fine-tuning layers (Ego2Exo).}}
\label{tab:tttlayers_ego}
\begin{tabular}{c c}
\toprule
\textbf{Layers} & \textbf{Ego-IoU}$\uparrow$  \\
\midrule
3  & 41.90 \\
\rowcolor{cyan!10}
4 & \textbf{41.95} \\
5   & 41.94  \\
6   & 41.93 \\
7   &  41.87 \\
\bottomrule
\end{tabular}

\end{minipage}
\hfill
\begin{minipage}{0.45\linewidth}
\centering
\caption{{\bf Ablation of fine-tuning layers (Exo2Ego).}}
\label{tab:tttlayers_exo}
\begin{tabular}{c c}
\toprule
\textbf{Layers} & \textbf{Exo-IoU}$\uparrow$  \\
\midrule
8  & 47.06 \\
9 & 47.14 \\
10   & 47.17 \\
\rowcolor{cyan!10}
11  & \textbf{47.18} \\
12  & 47.14 \\
\bottomrule
\end{tabular}
\end{minipage}

\end{table}

\section{Effeciency Analysis}
\label{sec:efficiency}
\begin{figure}
\centering
\includegraphics[width=0.48\textwidth]{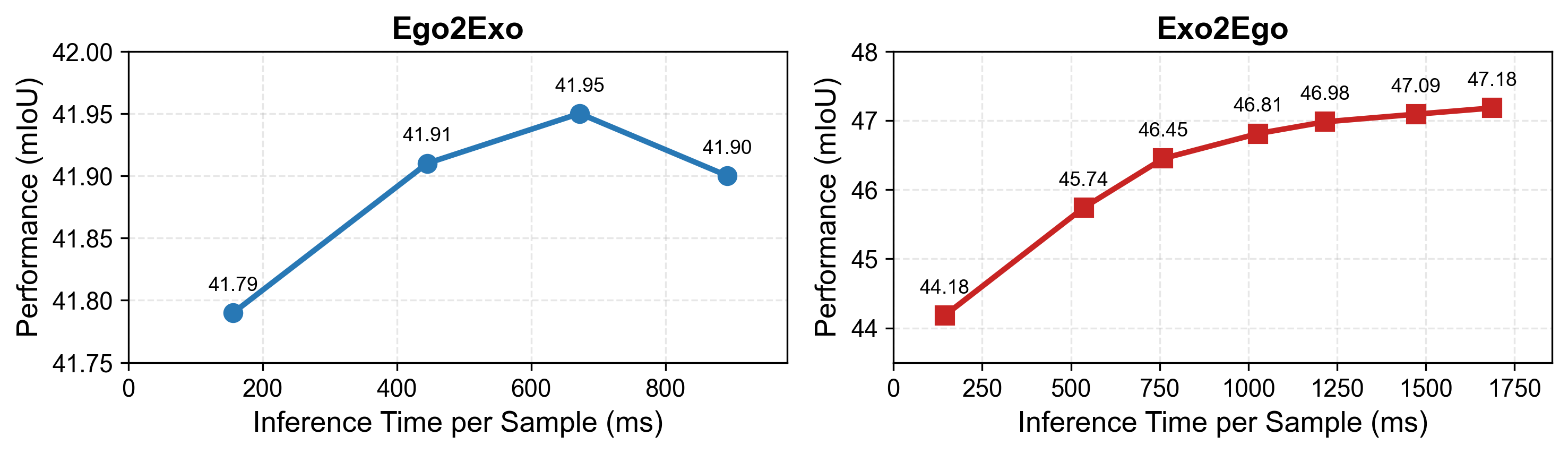}
\caption{{\textbf{Performance–latency trade-off under test-time training.} } 
}
\label{fig:efficiency}
\end{figure}

We agree that performance--latency trade-offs better reflect practical deployment than a single inference-time number. Accordingly, we provide an efficiency analysis in Figure~\ref{fig:efficiency}, reporting mIoU as a function of inference time by varying the number of test-time optimization steps (from 0 to 1, 2, and beyond). As shown, most of the performance gain is achieved with only 2 gradient updates, while further updates bring diminishing returns. This indicates that our method can achieve improvement with limited additional latency.

\section{More Qualitative Results}
\label{sec:more_vis}
\begin{figure*}
\centering
\includegraphics[width=1\textwidth]{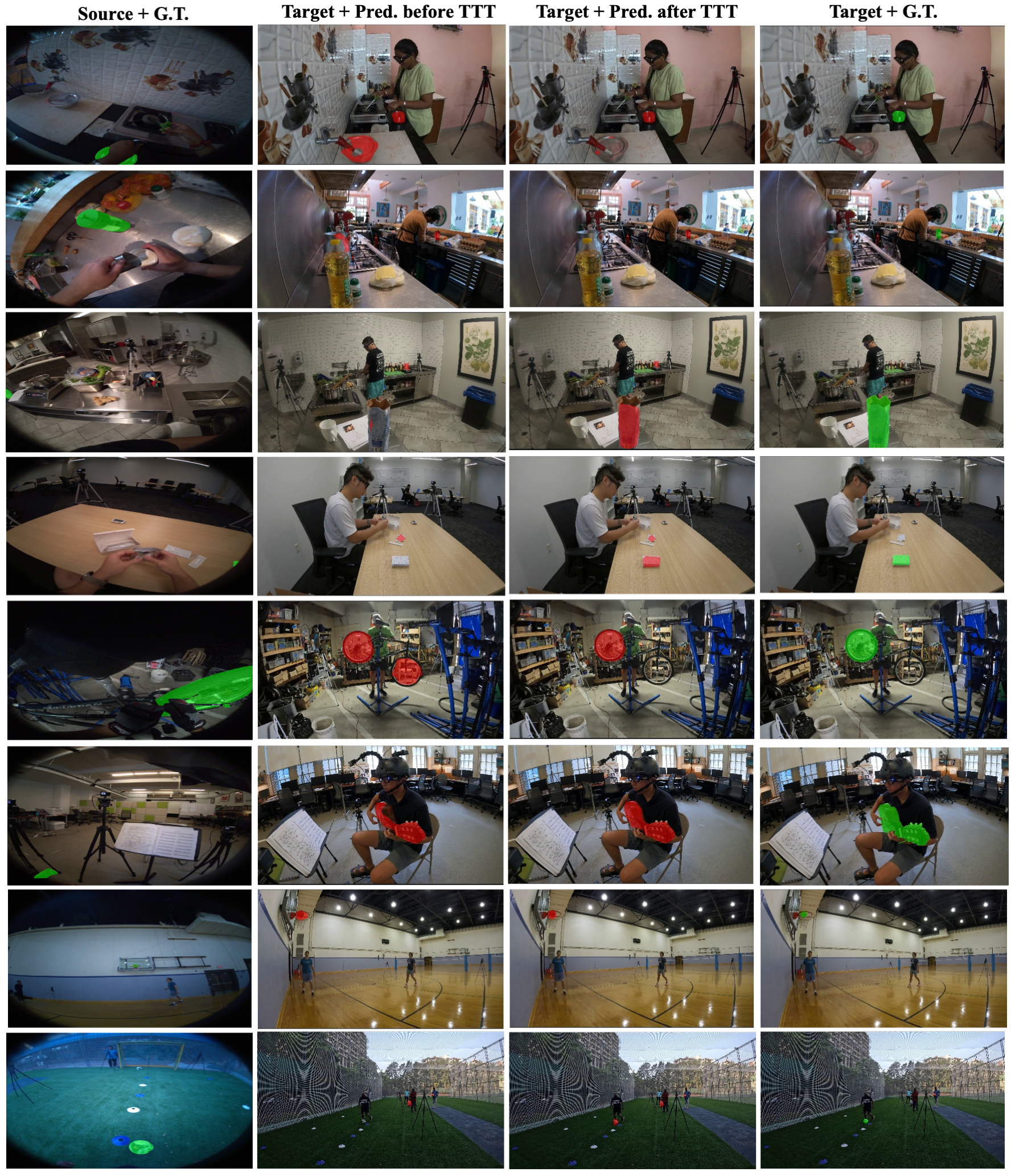}
\caption{
{\bf More qualitative results on the Ego-Exo4D correspondence benchmark (Ego2Exo).}
Each row shows a representative sample. Rows 1–3 correspond to cooking, row 4 to health, row 5 to bike repair, row 6 to music performance, row 7 to basketball, and row 8 to soccer.
}
\label{fig:app_qualitative_ego}
\end{figure*}

\begin{figure*}
\centering
\includegraphics[width=1\textwidth]{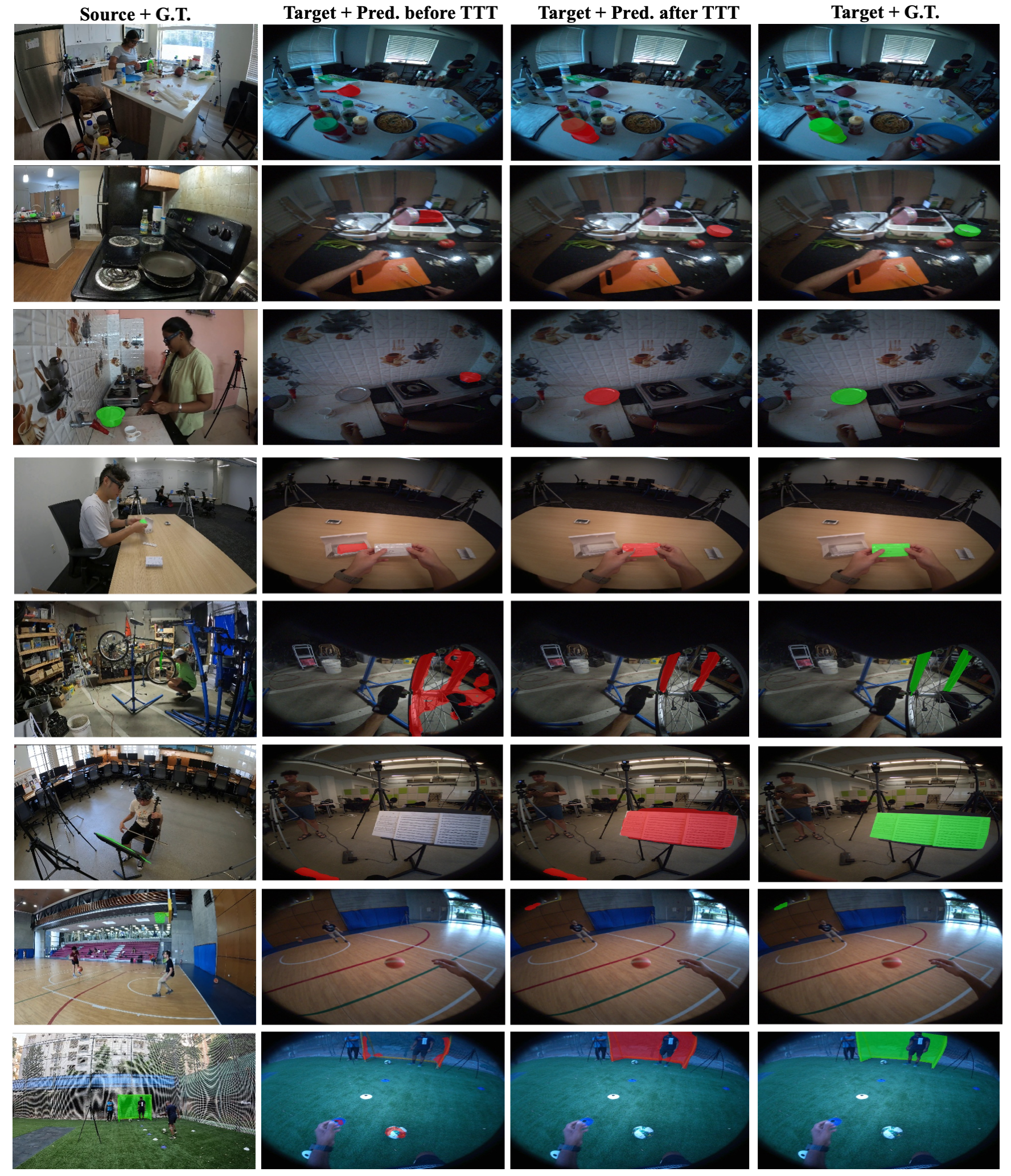}
\caption{
{\bf More qualitative results on the Ego-Exo4D correspondence benchmark (Exo2Ego).}
Each row shows a representative sample. Rows 1–3 correspond to cooking, row 4 to health, row 5 to bike repair, row 6 to music performance, row 7 to basketball, and row 8 to soccer.
}
\label{fig:app_qualitative_exo}
\end{figure*}

We provide additional qualitative results on the Ego-Exo4D correspondence benchmark in Figure~\ref{fig:app_qualitative_ego} and Figure~\ref{fig:app_qualitative_exo}. We present diverse examples that cover all six scenarios: cooking, health, bike repair, music, basketball, and soccer. The cooking scenario occupies three rows in each figure because it constitutes the largest portion of the benchmark. Across all scenarios, our method consistently produces masks that closely match the ground truth annotations, demonstrating strong robustness to variations in scene context, object category, occlusion, and viewpoint. These results also illustrate the effectiveness of TTT, which enables the model to focus more accurately on the target object while suppressing visually similar distractors and to generate masks that more completely cover the ground truth regions.

\begin{figure*}
\centering
\vspace{0.05in}
\includegraphics[width=1\textwidth]{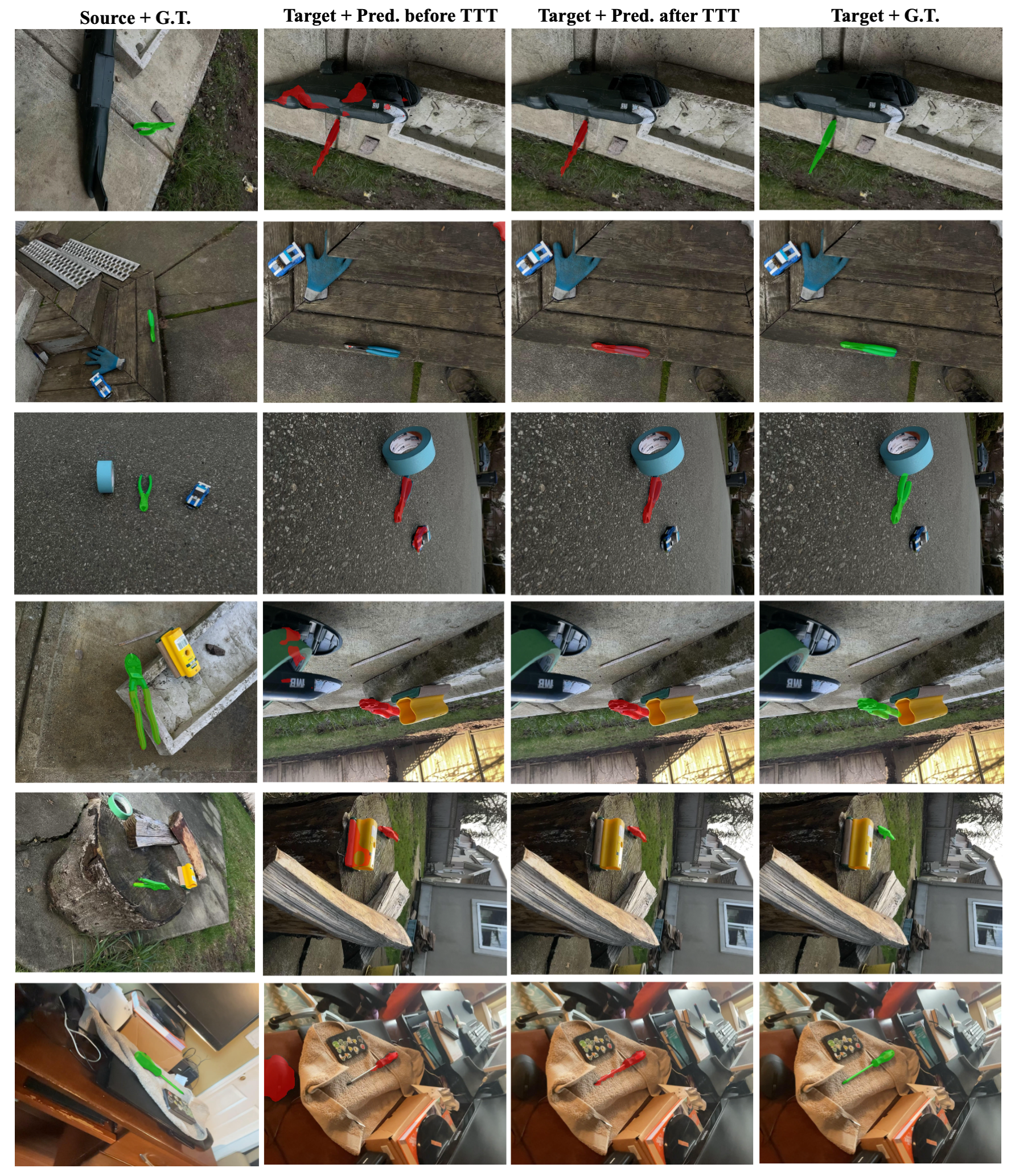}
\caption{
{\bf More qualitative results on the HANDAL-X benchmark.}
Each row shows a representative sample. 
}
\label{fig:app_qualitative_handalx}
\vspace{0.05in}
\end{figure*}

We further present qualitative results on HANDAL-X in Figure~\ref{fig:app_qualitative_handalx}, illustrating six examples spanning diverse hand–object interaction categories and highlighting the effectiveness of TTT. Our method successfully recovers the target masks in most cases.

\section{Limitation and Future Work}
\label{sec:limit}
All qualitative results are presented without excluding failure cases, providing a comprehensive view of the model’s potential errors. We summarize the common failure patterns, from frequent to rare:

\begin{itemize}[leftmargin=2em]
\item Incomplete coverage of the ground-truth regions.
\item Attraction to objects visually similar to the target object in the scene.
\item Complete failure to detect the target object.
\end{itemize}

We observe that TTT partially mitigates these errors, though some failures persist, leaving room for further improvement.

For future work, we plan to incorporate temporal cues to better capture object dynamics and further reduce the failure patterns identified above.

\end{document}